\pdfoutput=1
\documentclass[11pt]{article}

\usepackage{coling}

\usepackage{times}
\usepackage{latexsym}

\usepackage[T1]{fontenc}
\usepackage[utf8]{inputenc}

\usepackage{microtype}

\usepackage{inconsolata}

\usepackage{graphicx}
\usepackage{amsmath}
\usepackage{amssymb}
\usepackage{array}
\usepackage{multirow}
\usepackage{xcolor}
\usepackage{tikz}
\usepackage{newtxmath}
\usepackage{dsfont}
\usepackage{tabularx}
\usepackage{authblk}

\title{PMoL: Parameter Efficient MoE for Preference Mixing of LLM Alignment}

\author[a]{Dongxu Liu}
\author[a]{Bing Xu \thanks{Corresponding author: hitxb@hit.edu.cn}}
\author[c]{Yinzhuo Chen}
\author[a]{Bufan Xu}
\author[b]{Wenpeng Lu}
\author[a]{Muyun Yang}
\author[a]{Tiejun Zhao}
\affil[a]{Harbin Institute of Technology}
\affil[b]{Qilu University of Technology}
\affil[c]{Beijing Normal University - Hong Kong Baptist University United International College}

\begin{document}

\definecolor{mycolor1}{RGB}{195,248,208}
\definecolor{mycolor2}{RGB}{234,159,159}
\definecolor{mycolor3}{RGB}{248,244,127}
\definecolor{mycolor4}{RGB}{131,218,250}
\definecolor{mycolor5}{RGB}{154,120,231}
\definecolor{mycolor6}{RGB}{75,75,75}

\maketitle
\begin{abstract}

Reinforcement Learning from Human Feedback (RLHF) has been proven to be an effective method for preference alignment of large language models (LLMs) and is widely used in the post-training process of LLMs. However, RLHF struggles with handling multiple competing preferences. This leads to a decrease in the alignment of LLMs with human preferences. To address this issue, we propose Preference Mixture of LoRAs (PMoL) from the perspective of model architecture, which can adapt to any number of preferences to mix. PMoL combines Mixture of Experts (MoE) and Low Rank Adaptor (LoRA). This architecture is innovatively applied to the research of preference alignment and has achieved significant performance improvement. The expert group soft loss is used to enable MoE with the ability to mix preferences. Through comprehensive evaluation by the reward model and GPT-4o, the experiment results show that PMoL has superior preference mixing capabilities compared to baseline methods. PMoL achieves better preference alignment with lower training costs.

\end{abstract}

\section{Introduction}

% These instructions are for authors submitting papers to the COLING 2025 conference using \LaTeX. They are not self-contained. All authors must follow the general instructions for COLING 2025 proceedings which are an adaptation of (or rely on) the general instructions for ACL proceedings\footnote{\url{http://acl-org.github.io/ACLPUB/formatting.html}}, and this document contains additional instructions for the \LaTeX{} style files.

% The templates include the \LaTeX{} source of this document (\texttt{coling\_latex.tex}),
% the \LaTeX{} style file used to format it (\texttt{coling.sty}),
% a COLING bibliography style (\texttt{coling\_natbib.bst}),
% an example bibliography (\texttt{custom.bib}),
% and the bibliography for the ACL Anthology (\texttt{anthology.bib}).

% Large language models(LLMs) have demonstrated remarkable capabilities in many natural language processing tasks and multimodal tasks[]. An important prerequisite for using and offering such powerful LLMs to the public is that LLMs can generate behaviors that match human preferences, thus ensuring safe and controllable outputs []. Currently, a widely used alignment method is RLHF with the PPO algorithm, which first requires building a reward model using preference data, and then training the LLMs with reinforcement learning methods.

Currently, RLHF \citep{RLHF1,RLHF2}  is a widely used method to align LLMs with human preferences in terms of style and moral values.
This method requires the use of human-annotated preference data to build a reward model, and then uses reinforcement learning to train LLMs.

% RLHF leverages the broad knowledge and capabilities of LLMs to promote desired responses and behaviors, resulting in more harmless, helpful, and controllable AI systems. 
% % Both GPT-4 (OpenAI, 2023) and Anthropic (Ganguli et al., 2022) technical reports show that they used red-teaming during the RLHF phase to reduce potential harm of their models.
% Both GPT-4 (OpenAI, 2023) and Anthropic (Ganguli et al., 2022) technical reports show that they try to reduce potential harm and increase helpfulness of their models during the RLHF phase.
% However, in practice, the pursuit of increasing helpfulness and harmlessness often leads to contradictions (Ganguli et al., 2022; Bai et al., 2022a). For example, a model that refuses to answer can be considered safe but may also render the response unhelpful in extreme scenarios.
RLHF leverages the pre-trained knowledge and capabilities of LLMs to steer LLMs towards the direction desired by humans 
 \citep{pretrain1,pretrain2,pretrain3}, resulting in LLMs that are more harmless, helpful, and controllable.
Technical reports of LLMs like GPT-4 and Llama3 \citep{gpt4,llama3} show that they attempt to reduce potential harm and increase helpfulness of their models during the RLHF phase.
However, simultaneously pursuing increased helpfulness and harmlessness is contradictory \citep{HH1,HH2}. For example, refusing to answer is a harmless action, but always refusing to answer is unhelpful.
% Research [] has demonstrated that relying on a single reward, the RLHF method cannot satisfy diverse human preferences.
RLHF methods cannot satisfy diverse human preferences with a single reward \citep{maxmin}, but human preferences can also be further refined \citep{ultrafeedback}.
How to mitigate the competing between different preferences has become a worthwhile question to research.

\begin{figure}[t]
  \includegraphics[width=\columnwidth]{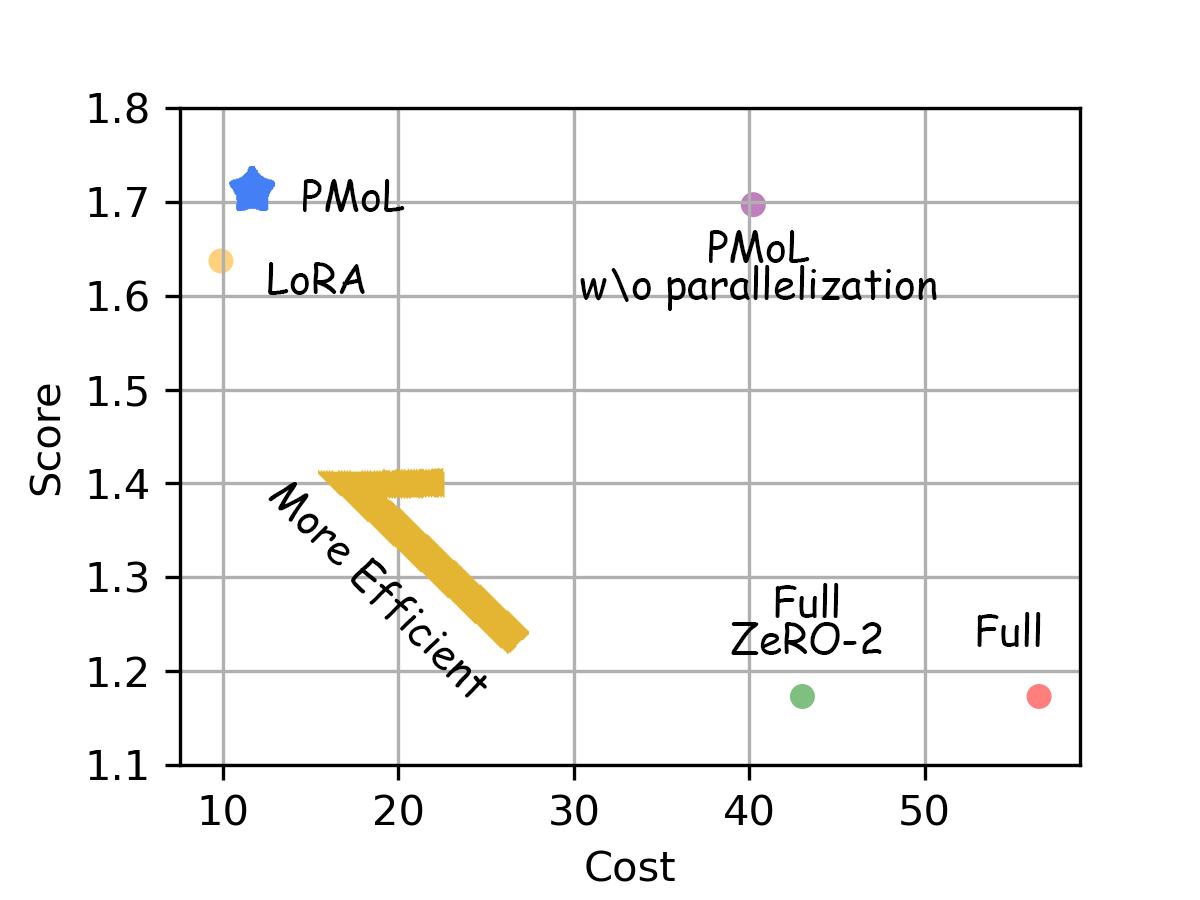}
  \caption{Results of training full parameters (denoted as Full, with and without ZeRO-2), LoRA, and PMoL on 30,000 pieces of data with 2 epochs. The horizontal axis represents the product of GPU memory usage (GB) times time (hours). The vertical axis represents the average score. The training uses 8 GPUs.}
  \label{fig:train cost}
\end{figure}

% In order to address this issue, Amini et al. [3] and Zhou et al. [53] propose the use of a score function to self-refine the objective. Cui et al. [16] and Tunstall et al. [43] employ GPT-4 [1] to select high-quality response pairs by scoring them.
% Research [] proposes learning multiple reward models and using dynamic mixing. Maxmin[] uses the EM algorithm to mix different preference rewards.
To address this issue, previous studies have tried training multiple reward models and dynamically mixing them \citep{multipul1,multipul2,multipul3,beavertails}, or controlling different preferences by synthesizing or filtering data \citep{ultrafeedback,multipul4}.
% However, these methods introduce a significant amount of additional training cost, making the costly RLHF method even more expensive. when the preference data is divided more finely, the cost grows exponentially. These methods are also less transferable.
% Therefore, we pose a question: Is there a method that can mix different preferences better at a lower cost?
However, these methods require stronger LLM supervision, additional data processing, or using more reward and advantage models. This introduces significant additional training costs, making the costly RLHF method even more expensive.
The cost increases exponentially as the number of preferences increases. 
% These methods are also less transferable.
These methods also have limited portability and extensibility.
Therefore, we consider a question: Is there a method that can mix multiple preferences better at a lower cost?

In this work, we introduce PMoL, which mixes different preferences from the perspective of model architecture. We find that the preference mixing task can be represented as a MoE framework. 
% Therefore, PMoL combines Parameter-Efficient Fine-Tuning (PEFT) with the MoE framework.
Therefore, PMoL integrates multiple LoRAs using the MoE framework.
By having different expert networks focus on different preference data, it mixes various preferences. PMoL coordinates all experts through a routing layer, determining each expert’s wight based on the context.

By introducing Expert Group Soft Loss, PMoL groups experts. 
% One group of experts focuses on one type of preference, while experts with different preferences are combined with each other.
This not only enhances PMoL's attention to each preference but also better utilizes the non-competing parts between preferences, promoting the capability of mixing multiple preferences. 
% Additionally, we find that the pre-trained knowledge is sufficient for the model to decode aligned with preferences. 
Additionally, Preference alignment do not perturb many tokens, and the aligned LLM still highly relies on pre-trained knowledge \citep{shift}. 
Therefore, we introduce the Empty Expert. The empty expert occupies some of the weights, allowing the model to decode only relying on pre-trained knowledge.

% Because the parallelization computation process, leading to lower training and inference cost compared to similar methods. 
Due to the parallelization of the computation process, PMoL has a lower training cost compared to similar methods.
As shown in Figure \ref{fig:train cost}, our method uses less GPU memory, less training time, and achieves better preference mixing results.
The empathy is added as the third preference in addition to helpfulness and harmlessness, using the HHRLHF \citep{HH2}, Extes \citep{extes}, and ESConv \citep{esconv} datasets combined to form a preference dataset. PMoL is validatd the superior performance in reducing training cost and mixing multiple preferences.

Our main contributions are as follows in three aspects:

\begin{itemize}
\item[1.] We design PMoL, a new method that integrates multiple LoRAs within MoE framework, allowing for efficient mixing of multiple preferences. PMoL has good versatility and not limited by the type of preferences or alignment algorithms.
% We design a new method that can efficiently mix multiple preferences. This method is not designed for any specific preference or alignment algorithm and plug-and-play, with good transferability and excellent performance.
% We design a new method that can efficiently mix multiple preferences. This method is not designed for any specific preference or alignment algorithm. It is plug-and-play, demonstrating good transfer ability. The effectiveness of this method -------.
\item[2.] The PMoL has low training cost. Compared to other methods, PMoL does not require loading additional models and has no extra processing for reward and data, allowing it to adapt to any number of preference mixing. 
% We parallelize the computation process, boosting training speed by 5 times and inference speed by 3 times.
% We achieve preference mixing with lower training cost. Compared to other methods, our method does not need to load additional models and does not have additional preference processing and mixing phases. The training cost is not significantly different from general PEFT methods.
% \item[3]  
% We parallelize the computation process, which significantly reduce the training and inference costs compared to other similar methods.The training and inferencce cost is not significantly different from general PEFT methods.
% We parallelize the computation process, which significantly reduce the training and inference costs compared to other similar methods.
\item[3.] We consider sentiment as preference and construct a multi-preference dataset that includes empathy preference. 
% It is proven that empathy preference also compete with helpfulness and harmlessness.
It is proven that considering empathy preference as a separate type of preference is necessary.

\end{itemize}
\section{Relative Work}
\subsection{PEFT and MoE}
Parameter-Efficient Fine-Tuning (PEFT) freezes the LLM backbone and introduces a small number of additional parameters for finetuning. It can be divided into two categories: one is the adapter-based methods, which adds a trainable networks to the Transformer layers, such as LoRA, IA3 \citep{ia3,lora}. The other is the prompt-based method, which adds continuous or discrete trainable vectors to the input, such as P-tuning \citep{prompttune1}.
% ,prompttune2,prompttune3
Mixture-of-Experts (MoE) uses multiple sparse, activatable expert networks to replace the feedforward neural networks, which significantly enlarges the model without remarkably increasing the computational cost \citep{moe}. Currently, the Token-level routing MoE architecture has been widely applied in LLMs \citep{moetoken1,moetoken2,moetoken3,moetoken4}.

Some studies have improved model performance by using multiple LoRAs. \citet{loramoe1} proposes using multiple LoRA adapters to achieve multi-task generalization. \citet{loramoe2} uses thousands of LoRAs to provide services. \citet{loramoe3,loramoe4,loramoe5} combine MoE and PEFT module and applies it to LLM finetuning, achieving performance close to or even exceeding supervised fine-tuning (SFT) on specific downstream tasks. PEMT \citep{loramoe6} integrates multiple pre-trained LoRAs in the form of MoE.

These methods primarily focus on the combination of MoE and PEFT to simple downstream tasks with SFT. However, there is no research on how to apply these methods to complex tasks such as preference alignment.

\begin{figure*}[htbp]
  \centering
  \includegraphics[width=0.96\linewidth]{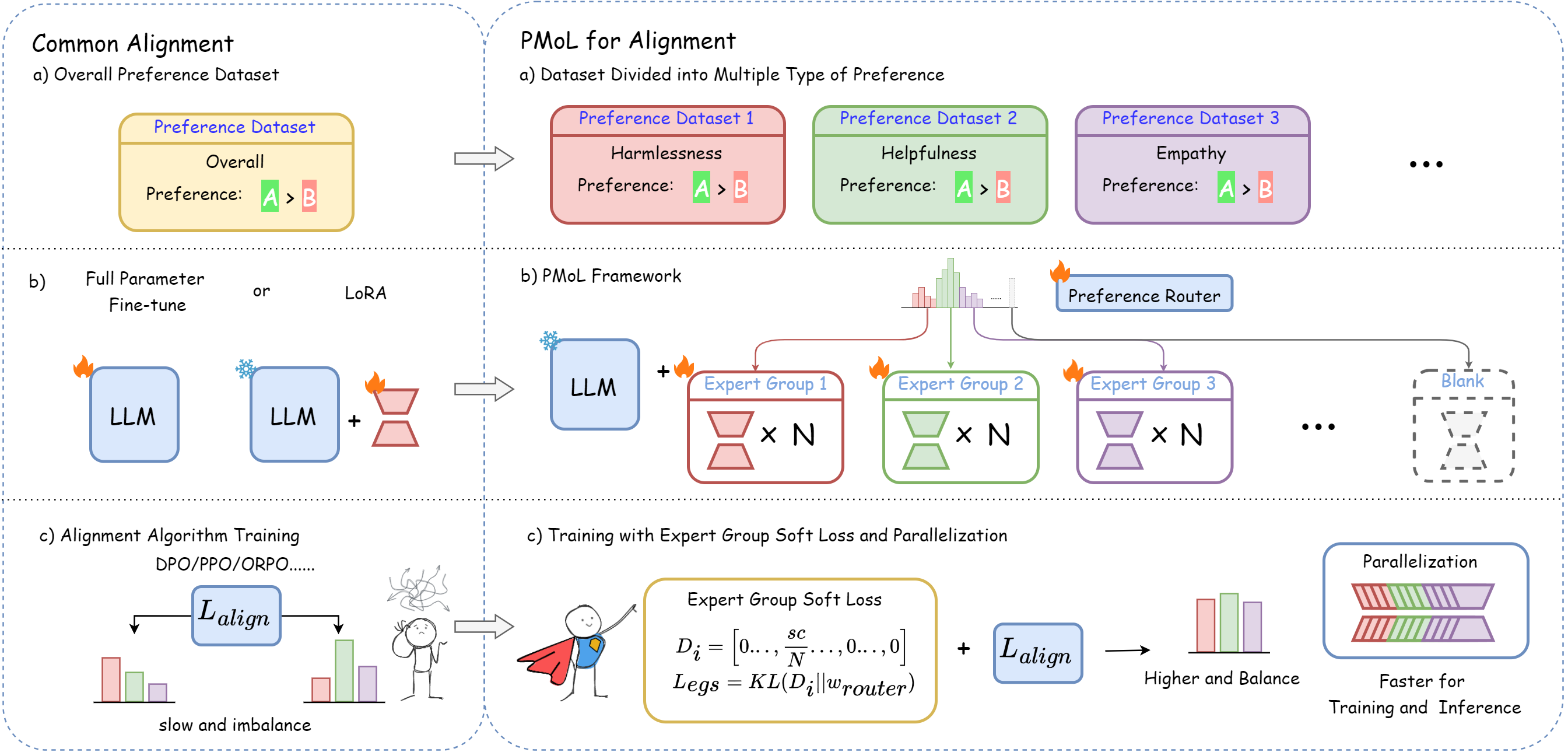} 
  \caption {The overview of PMoL. PMoL is an adaptor with MoE framework added to the Transformer block. All experts are divided into multiple groups.  Each group of experts focuses on one preference. Additional empty expert allows the LLM to reduce its reliance on experts. During the training phase, the pre-trained parameters are frozen, and only the LoRA experts and Router are updated.}
  \label{fig:overview}
\end{figure*}

\subsection{Preference Alignment and Mixing}
Current RLHF methods involve training a reward model based on human preference, and then fine-tuning the language model using Proximal Policy Optimization (PPO) \citep{ppo}.
Due to the high cost of PPO training, methods like DPO \citep{dpo} that do not require a reward model have emerged. Many variants of DPO methods, such as ORPO \citep{orpo}, have improved DPO.

% These methods train on data without distinction, ignoring the diversity and potential conflicts in human preferences. RLHF[] points out that mixing helpfulness and harmlessness preference data can not lead to the performance as well as trained on single preference data, indicating a conflict between the two types of preferences.

Standard RLHF does not effectively mix multiple preferences \citep{RLHF2}. 
To address this issue, some studies have focused on data, calibrating the reward model to be consistent with the data \citep{mixing1,mixing2}. Other studies emphasize the shortcomings of a single reward model, training reward models to learn mixed preferences \citep{maxmin} or training reward model for each type of data and mixing them \citep{safe}.

These methods mainly focus on data and reward models, which result in high training or manual cost. Most of them are only applicable to two types of preferences. When more preference categories are introduced, the cost grows exponentially or method becomes unavailable.

% \section{Preliminary}

% \subsection{Preliminary Experiment}
% \label{sec:Distribution_Difference}

% Research[] proposes that between the aligned and base models, the distribution shift of the next token is not significant. The shift becomes smaller during decoding.
% We replicate the experiment.
% As shown in Figure \ref{fig:shift},\ref{fig:shift2}, we do not observe a trend of the KL divergence decreasing. There are only a few positions where the KL divergence value is large, indicating a significant distribution shift.

% % We color the words according to the distribution shift. The results are shown in Figure \ref{fig:shift example}. At most positions, the token with the highest probability decoded by the aligned and base model is the same one.

% These experiments show that, 
% % in most position in decoding, base model can complete the same result as the aligned model relying on pre-trained knowledge. 
% % Preference alignment can be achieved by perturbing the distribution of only a few tokens.
% the preference alignment do not perturb many tokens, and the aligned LLM still highly relies on pre-trained knowledge.
% We present more experimental details and findings in Appendix \ref{append:Preliminary}.

\section{PMoL}
The overview of PMoL is shown in Figure \ref{fig:overview}. PMoL is an adaptor with MoE framework using multiple PEFT modules as experts added in transformer block.
Experts are grouped and an additional loss function is uesd to restrict one group of experts to focus on one type of preference. Router determines the mixing ratio of all experts based on the context information. The empty expert enables LLM to utilize pre-trained knowledge more efficiently.
\subsection{Preference Mixing as MoE}
\label{sec:Why_MoE_can_Mix_Preference}
Firstly, The process of LLM decoding the next token can be represented as
% \begin{equation}
%   \label{eq:example}
%   \begin{aligned}
%     & Response[n] = \\
%     M_{P}(I & nstruct+Response[:n]) 
%   \end{aligned}
% \end{equation}
\begin{equation}
  \label{eq:moti1}
    Res[n] = M_{P}(Ins+Res \left [:n \right ]) 
\end{equation}
Where $P$ represents the parameters of LLM $M$. Ins represents instruction, and Res represents the LLM's response.

The parameter difference between the LLM before and after preference learning is denoted as $\Delta P$. The process of the LLM decoding the next token after preference learning can be represented as
\begin{equation}
  \label{eq:moti2}
  Res[n] = M_{P+\Delta P}(Ins+Res[:n])
\end{equation}

% As described in section \ref{sec:Distribution_Difference}, 
% At most position in decoding, base model can complete the same result as the aligned model relying on pre-trained knowledge.[]
Preference alignment can be achieved by perturbing the distribution of only a small number of tokens \citep{shift}.
Therefore, we set a coefficient $w$ for the parameter difference $\Delta P$, which allows LLM to adjust the degree of participation during the decoding process and rely more on pre-trained knowledge. The value of $w$ positively correlate with the distribution difference between the aligned and base LLM. The decoding process can then be represented as:
\begin{equation}
  \label{eq:moti3}
Res[n] = M_{P+ w \Delta P} (Ins+Res[:n])
\end{equation}
where $w$ denotes the coefficient of the parameter difference $\Delta P$, which is determined by the context.

Finally, The parameter difference $\Delta P$ is replcaed by multiple parameter differences of preferences. 
Each of parameter differences is assigned a coefficient.
% Due to the superficial alignment hypothesis [], we allow LLM to decode based on pre-trained knowledge alone, meaning the sum of all expert coefficients can be less than 1.
The coefficient $w$ only represents the participation degree of preference, so the sum of coefficients is not restricted to be 1.
% The sum of all expert coefficients is allowed to be less than 1. 
The decoding process can be represented as
\begin{equation}
  \label{eq:moti4}
  Res[n] = M_{P+\Sigma w_i \Delta P_{i}} (Ins+Res[:n]) 
\end{equation}

We denote each parameter difference $\Delta P$ as an expert $E$, 
% and restrict the sum of the coefficients, 
this decoding process can be equated to the MoE form.
% In this paper, to reduce training cost, we select LoRA as the experts, modeling the parameter differences for each preference.

% \section{MoLE}

\subsection{LoRA as expert}
Since the parameter differences for each preference need to be represented through the PEFT module, LoRA is chosen as experts in MoE. 
For a linear layer with parameter $W$, LoRA introduces two lower-rank matrices, 
$ A \in \mathbb{R}^{a\times r} $ and 
$ B \in \mathbb{R}^{r\times b} $ 
,where $r\ll\min(a,b)$ . The output of linear layer with LoRA added can be represented as 
\begin{equation}
  \label{eq:method1}
  h=W_0x+BAx
\end{equation}

\subsection{Mixture of LoRA}

As described in section \ref{sec:Why_MoE_can_Mix_Preference}, The MoE framework is uesd to mix different preferences. Different experts need to be coordinated through a router. The output of the MoE module can be expressed as: 
\begin{equation}
  \label{eq:method2}
  h=W_0x+\Sigma_{i=0}^{K} w_{router}[i] E_i(x)
\end{equation}
where each expert $E_i$ is a LoRA, $K$ is the total number of experts, and $w_{router}[i]$ is the weight of each expert, which is obtained through router. The calculation process of the router is expressed as:
\begin{equation}
  \label{eq:method3}
  w_{router}=softmax(W_rx+b)
\end{equation}
% \begin{equation}
%   \label{eq:method3.1}
%   w_i=softmax(w_o)[i]
% \end{equation}
where $ W_r \in \mathbb{R}^{h\times (K+1)} $ is a trainable matrix that can assign weights to each expert based on the context vector x. 

% As described in Equation \ref{eq:moti4}, we allow the sum of all expert weights to be less than 1.
To ensure both stability and flexibility, Equation \ref{eq:moti4} is modified to restrict the sum of all weights to be less than 1.
For this reason, the empty expert is set. The number of router’s output is set to K+1, with K weights assigned to the K experts and the remaining weight representing the empty expert. The empty expert allows the model to utilize pre-training knowledge more efficiently.

\subsection{Expert Group Soft Loss}
Experts utilization imbalance is a classical problem in MoE. In this work, we need to ensure a balanced experts utilization while limiting different expert to focus on different preferences.

For the $K$ preferences need to be mixed, each preference corresponds to a dataset $D_i$, and the overall data is $ D = \{ D_1,D_2 ... D_K \} $.  
The correspondence between the preference dataset $D_i$ and the experts is represented as $ T_i = ( h_i,s_i)$, where $h_i$ and $s_i$ denote the start and end indices of the expert range.

In training, for a preference data from dataset $D_i$, we need to compute the standard balance distribution $D_i$ firstly:
\begin{equation}
  \label{eq:method4}
  \begin{aligned}
  D_i & = \{0\} \times K \\
sc_d & = sc / (s_i-h_i) \\ 
D_i[h_i:s_i] & = sc_d  \\
  \end{aligned}
\end{equation}
where $sc$ is the soft constraint coefficient. Different soft constraint coefficients can be assigned to different preference dataset to control the mixing degree

Expert group soft loss is the KL scatter between the expert weights output by router and the standard balance distribution:
\begin{equation}
  \label{eq:method5}
  L_{egs} = KL(D_i||w_{router}[:K])
\end{equation}
Overall, $L_{egs}$ ensures that experts in different groups focus on different preference data, and the experts utilization in the same group is balanced. The total loss is expressed as:
\begin{equation}
  L = L_{DPO} + \beta L_{egs}
  \label{eq:beta}
\end{equation}
where $\beta$ is the coefficient that controls the strength of expert group soft loss. 

In training , the backbone of LLM is frozen and only the router parameters and LoRA expert is trained. During inference, the router automatically activates the corresponding expert based on the context without specifying the data type. Note that the empty expert does not participate in the computation of the expert group soft loss.

\subsection{Parallelization}
We analyze the time consumption of each part of PMoL during training and inference. As shown in Table \ref{tab:time}, it is found that the time consumed to compute PMoL is too long. 
% The GPU utilization is also low. 
Therefore, PMoL seriously slows down the training and inference of the model.
Although $L_{egs}$ is time-consuming, it is executed infrequently and has minimal impact.
As shown in Table \ref{tab:time}, the parallelized computation process significantly reduces the time consumption in the training and inference process.

\begin{table}[htbp]
  \centering
  \begin{tabular}{lcccc} % 定义表格的列格式：左对齐 | 居中对齐
    \hline
    \multirow{2}*{\textbf{step}} & \multicolumn{2}{c}{\textbf{Parallelized}} & \multicolumn{2}{c}{\textbf{Unparallelized}}\\ 
    % \hline
     & train & inference & train & inference\\ 
    \hline
    PMoL & 0.251 & 0.229 & 3.310 & 3.297 \\
    Linear & 0.057 & 0.046 & 0.057 & 0.046\\
    $L_{dpo}$ & 0.308 & \_ & 0.310 & \_ \\
    $L_{egs}$ & 17.24 & \_ & 17.28 & \_ \\
    \hline
  \end{tabular}
  \caption{The time consumption for each part calculates once during training and inference in milliseconds.}
  \label{tab:time}
\end{table}

\section{Experiments}

\begin{table*}[htbp]
  \centering
  \setlength{\tabcolsep}{4pt}
  \begin{tabular}{lccccccccccc}
    \hline
    \multirow{2}*{\textbf{method}} & \multirow{2}*{\textbf{preference}} & \multicolumn{2}{c}{\textbf{helpfulness}} & \multicolumn{2}{c}{\textbf{harmless}} & \multicolumn{2}{c}{\textbf{empathy}} & \multicolumn{2}{c}{\textbf{average}}& \multicolumn{2}{c}{\textbf{$\Delta$}}\\
    & & rw & LLM & rw & LLM & rw & LLM & rw & LLM & rw & LLM \\ 
    \hline
     Full & \multirow{5}*{All} & 0.875 & 2.561 & \textbf{0.982} & 1.983 & 1.572 & 3.589 & 1.137 & 2.711 & \_ & \_\\
     LoRA & & \textbf{1.474} & 2.854 &  0.615 & 1.867 & \textbf{2.762} & \textbf{4.542} & 1.637 & 3.087 & 0.280 & 0.442 \\
     PMoL & & 1.447  & \textbf{2.956} & 0.962 & \textbf{2.004} & 2.735 & 4.538 & \textbf{1.714} & \textbf{3.166} & \textbf{0.224} & \textbf{0.376}\\ 
     JANUS & & 1.462  & 2.872 & 0.858 & 1.842 & 2.642 & 4.079 & 1.642 & 2.931 & \_ & \_\\
     JANUS* & & 1.426  & 2.503 & 0.720 & 1.661 & 2.354 & 3.863 & 1.500 & 2.675 & \_ & \_\\
     \hline
     Full  & \multirow{3}*{HH} & 0.972 & 2.763 & \textbf{0.994} & 2.039 & \_ & \_& 0.983 & 2.401 & \_ & \_\\
     LoRA  & & 1.492 & 2.981 &  0.684 & 1.439 & \_ & \_& 1.108 & 2.210 & 0.301 & 0.706\\
     PMoL  & & \textbf{1.507}  & \textbf{3.068} & 0.978 & \textbf{2.073} & \_ & \_& \textbf{1.242} & \textbf{2.571} & \textbf{0.213} & \textbf{0.430}\\
     \hline
     % LoRA & \multirow{2}*{helpfulness} & 1.597 & 3.204& \_ & \_& \_ & \_& \_ & \_& \_ & \_\\
     % PMoL & & 1.539 & 3.198 & \_ & \_& \_ & \_& \_ & \_& \_ & \_\\
     % \hline
     % LoRA & \multirow{2}*{harmless} & \_ & \_& 1.238 & 2.628& \_ & \_& \_ & \_& \_ & \_\\
     % PMoL & & \_ & \_& 1.372 & 2.804& \_ & \_& \_ & \_& \_ & \_\\
     % \hline
     % LoRA & \multirow{2}*{sentiment} & \_ & \_& \_ & \_& 2.857 & 4.757& \_ & \_& \_ & \_ \\
     % PMoL & & \_ & \_& \_ & \_& 2.903 & 4.625& \_ & \_& \_ & \_ \\
     LoRA & \multirow{2}*{Single} & 1.597 & 3.204 & 1.238 & 2.628 & 2.857 & 4.757& 1.897 & 3.530 & 0 & 0 \\
     PMoL & & 1.539 & 3.198 & 1.372 & 2.804 & 2.903 & 4.625 & 1.938 & 3.542 & 0 & 0 \\
    \hline
  \end{tabular}
  \caption{\label{tab:main_result}
    Helpfulness, harmlessness, and empathy scores given by the reward models and GPT-4o. The best values for each score are marked in bold. "Average" refers to the average score for each preference, with higher values indicating better preference mixing ability. $\Delta$ refers to the average difference in scores before and after preference mixing, with higher values indicating larger loss due to preference mixing. ALL includes all three types of preference data. HH only includes preference data for helpfulness and harmlessness. "Single" contains only one type of preference data, and the preference type is the same as the column name, used to show the upper limit of mixed preference performance without loss. “rw” refers to the results evaluated using the reward model. “LLM” refers to the results evaluated using GPT-4o.
  }
\end{table*}

\subsection{Experimental Setup}

\subsubsection{Data}
We use HHRLHF \citep{RLHF2} as the training and test dataset for helpfulness and harmlessness preferences. For empathy preference, we combine Extes \citep{extes} and ESConv \citep{esconv} datasets. Since the Extes and ESConv only contain one response and do not include preference pairs, we use instruction fine-tuned gemma-2b to sample the rejected responses in the preference data, using the responses from the original dataset as the chosen data. We extract 10,000 samples from each type of preference data as the training set.

\subsubsection{Evaluation}
We use reward models for evaluation. We add a value head to train the reward model. During testing, we judge the alignment of the model with preferences by the average scores from the reward model. We train reward models separately for each preference.
We also use LLM for evaluation. By prompting GPT-4o, we score the responses. For responses that are rejected by GPT-4o due to harmful content, we use human evaluation.

To measure the capability of preferences mixing, we calculate the average preference score and score difference before and after preferences mixing. Specifically, we subtract the score of the model trained on multiple preference data from the score of the model trained on single preference data. The average score difference across multiple preferences is taken. A larger difference indicates a greater loss due to preference mixing. 
A lower average score indicates a poorer capability of the method to mix preferences.

\subsubsection{Baselines}

We select some methods that can be used without significant extra costs. 
% Training details of PMoL are in Appendix \ref{append:tarin1}.

Full Parameter Fine-tuning (Full): Training the model with the DPO algorithm. All parameters of the model are set to be trainable. 
% Since the DPO method does not require reward model and advantage model, the extra cost of this method is only the GPU memory occupied by the reference model.

LoRA \citep{lora}: We add LoRA to the down\_proj layers on all transformer modules in the LLM. 
% Compared to full parameter fine-tuning, LoRA requires fewer gradients and activation during training. 
To compare fairly, we use the best-performing hyperparameters.

JANUS \citep{1000}:JANUS sets different system prompts for different preference data and trains the model through instruction fine-tuning. 
% Since instruction fine-tuning does not require a reference model, the training cost of JANUS is lower.
Notably, unlike PMoL, JANUS needs to be given the preference type during inference. 
% JANUS does not have the ability to automatically distinguish preferences.
Therefore, for a fair comparison, we also calculate the score for JANUS without given preference type (denoted as JANUS*).

\subsection{Main Results}

% As shown in Table [], we train PMoL on one, two, and three types of preference datasets. Compared to PMoL trained on one type of preference, PMoL trained on three types of preference only see a decrease of only 0.1 in helpfulness, 0.1 in harmless, and 0.1 in sentiment scores on the reward model, with an average decrease of 0.1. 
% LoRA’s scores on the three type of preferences decrease 0.1 in helpfulness, 0.1 in harmless, and 0.1 in sentiment, with an average decrease of 0.1. 
% PMoL loses fewer scores. This shows PMoL’s outstanding ability in preferences mixing.
As shown in Table \ref{tab:main_result}, we train PMoL on one, two, and three types of preference datasets. 

\label{sec:main_result}
PMoL achieves the highest average score. Although LoRA's average score is higher than Full's, LoRA shows significant imbalance on the three preferences. LoRA gets the lowest harmlessness score, while Full gets the highest harmlessness score but has low helpfulness and empathy scores. Both of them contradict the purpose of preference mixing. PMoL achieves a harmlessness score close to Full's while maintaining helpfulness and empathy scores close to LoRA's. This shows that PMoL balances the three preferences well.

\citet{RLHF2} have shown that there is a trade-off in scores between helpfulness and harmlessness preferences. 
PMoL sacrifices a bit of helpfulness and empathy scores to achieve a significant increase in harmlessness score. Full sacrifices more helpfulness and emotional scores, but there was no noticeable improvement in harmlessness score.
This highlights PMoL's advantage in mixing multiple competing preferences.

Compared to PMoL trained on one type of preference data, the average of reward model scores of PMoL trained on three types of preference data decreases 0.224.
However, the average score of LoRA decreases 0.280.
PMoL loses fewer scores. This shows PMoL’s outstanding capability in preferences mixing.

Compared to Full and LoRA methods, when the number of preference types increases from two to three, PMoL’s average score on HH decreases less (-0.0038 vs -0.0055 and -0.0044). This shows that with an increase in the number of preference types, PMoL maintains a high ability to mix preferences. It proves that PMoL has good scalability. The Full and LoRA methods, however, do not adapt well to a more types of preference data.

JANUS gets results close to PMoL.
However, when the system prompts with preference types are replaced with the default system prompt, JANUS’s score is not as high as PMoL’s.
This shows that JANUS heavily relies on the preference type given by the system prompt. JANUS has a poorer ability to mix and distinguish preferences.

\subsection{Ablation Study}

\subsubsection{Effect of Expert Group Soft Loss}

% With the expert balance loss during training encourages a group of experts to focus more on a certain preference. 
The average score after adding the expert group soft loss in Table \ref{tab:beta} shows its significant improvement on the preference mixing ability. 
Moreover, we analyze the effect of beta in Equation \ref{eq:beta}. As shown in Table \ref{tab:beta}, a larger $\beta$ balances different experts well but also restricts the combination of experts with different preferences. A smaller beta has difficulty in constraint, leading to a decrease in PMoL's performance.

In addition to Equation \ref{eq:beta}, we test some loss functions for comparison, including Switch Transformer \citep{switch}, which is commonly used in MoE.
% The detailed calculation process for all loss functions can be found in Appendix []. 
The detailed formulas can be found in Appendix \ref{append:loss}.
As shown in Table \ref{tab:loss}. Apart from soft and hard, the other loss functions do not have the ability to group experts. Therefore, their preference mixing ability is poor. Compared to soft, hard has stricter requirements of the expert weights. This can result in information that is not conflicting between multiple preferences not being shared by all experts.

\subsubsection{Effect of Empty Expert}

The effect of the empty expert is shown in Table \ref{tab:empty}. The empty expert provides some improvement to the preference mixing ability of PMoL. 
 
% However, the situation described in Section \ref{sec:Distribution_Difference} does not occur. 
As shown in Figure \ref{fig:activation}, the weights of the empty expert are below 0.1 for all preferences, indicating a minor effect. This is due to PMoL’s small perturbation to the model output. Compared to the linear layer, the average output of PMoL is 3 orders of magnitude smaller. The low-rank nature of LoRA means it does not cause significant perturbation to the main network.

\subsubsection{Effect of LoRA rank and number}

We adjust the rank and number of LoRA experts. 
As shown in Figure \ref{fig:lora abla} left, a smaller rank leads to poor fitting capability of the LoRA experts, failing to learn complex semantic information in a preference data. This phenomenon is particularly evident in the helpfulness and harmlessness preferences. When the rank exceeds 8, a larger rank does not bring significant improvement.

As shown in Figure \ref{fig:lora abla} right, a smaller number of experts causes the model to degrade into a structure similar to LoRA, reducing the mixing ability. When there are too many experts (128), we found that a large number of experts receive a very small weight (<0.001), and the scores also showed a significant decrease.

\section{Analyse}

\subsection{Parallelization}

% \begin{figure*}[htbp]
%   \includegraphics[width=0.48\linewidth]{figure/figure5.1.png} \hfill
%   \includegraphics[width=0.48\linewidth]{figure/figure5.2.png}
%   \caption {The training costs (left) and the cost of generating 100000 tokens (right) for full parameter fine-tuning, LoRA, PMoL w and w/o parallelization. The horizontal axis represents memory usage(GB) multiplied by time(Hour), and the vertical axis is the average score on three preferences. The training uses 8xA6000 GPUs.}
%   \label{palla}
% \end{figure*}

% \begin{figure*}[htbp]
%   \includegraphics[width=0.48\linewidth]{figure/figure5.1.png} \hfill
%   \includegraphics[width=0.48\linewidth]{figure/figure5.2.png}
%   \caption {The training costs (left) and the cost of generating 100000 tokens (right) for full parameter fine-tuning, LoRA, PMoL w and w/o parallelization. The horizontal axis represents memory usage(GB) multiplied by time(Hour), and the vertical axis is the average score on three preferences. The training uses 8xA6000 GPUs.}
%   \label{palla}
% \end{figure*}

\begin{figure}[t]
  \includegraphics[width=\columnwidth]{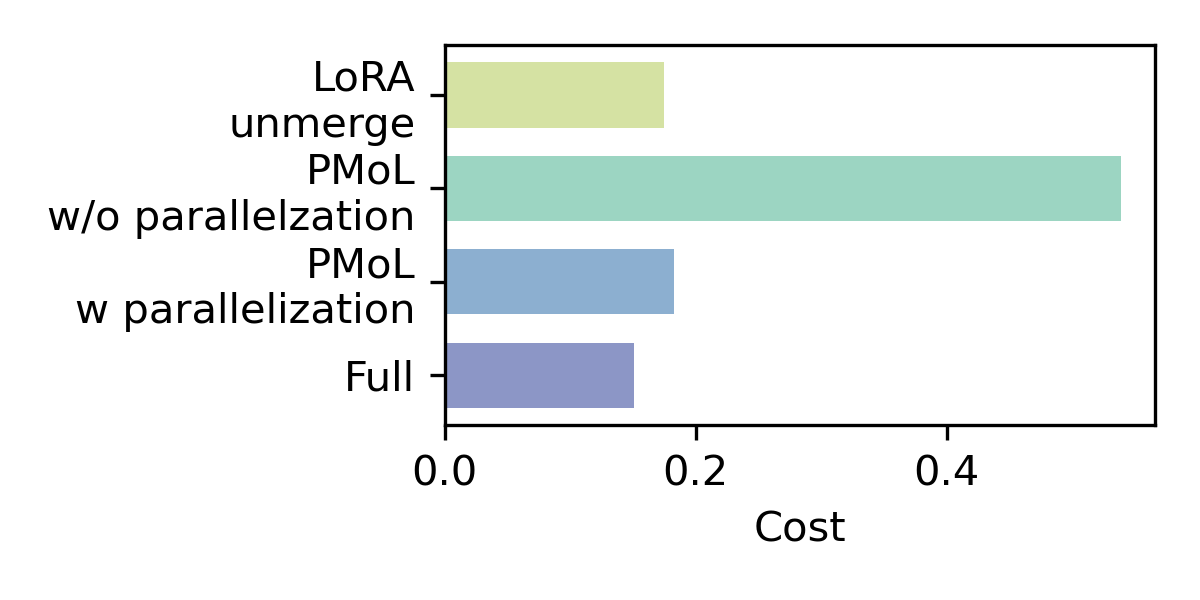}
  \caption{The cost of generating 100000 tokens for full parameter fine-tuning, LoRA, PMoL w and w/o parallelization. The horizontal axis represents memory usage(GB) multiplied by time(Hour).}
  \label{fig:palla}
\end{figure}

Figure \ref{fig:palla} and \ref{fig:train cost} show the significant advantage of parallelization. Compared to PMoL without parallelization, parallelization boosts training speed by 4 times and inference speed by 3 times. Parallelization achieves training cost and inference speed close to LoRA while also surpassing the preference mixing ablitity of full parameter fine-tuning.

\subsection{Conflict among Preferences}

As shown in Table \ref{tab:main_result}, whether with three or two preferences mixed, PMoL scores lower than when trained on a single preference.
A similar situation occurs in the results of the LoRA and Full methods.
This shows that there is a competing among the three preferences. The model needs to balance these preferences. The examples in Table \ref{tab:empathy example} further illustrate the competition between empathy, helpfulness, and harmlessness. This shows the rationality and necessity of considering empathy as a preference.
% However, PMoL gets a higher average score compared to LoRA and Full. This shows that PMoL can better balance multiple competing preferences.

\subsection{Impact of Data Gap on Preference Mixing}
 % on the Difficulty of Preference Mixing
% \begin{tikzpicture}
%   \fill[mycolor] (0,0) rectangle (0.2,0.2);
% \end{tikzpicture}

\begin{figure}[t]
  \includegraphics[width=\columnwidth]{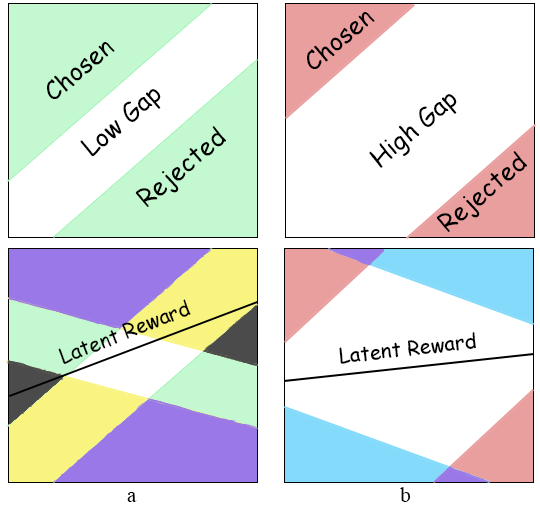}
  \caption{The difficulty of mixing preferences with different Gap data. 
   \begin{tikzpicture}
      \fill[mycolor1] (0,0) rectangle (0.2,0.2);
    \end{tikzpicture}
   \begin{tikzpicture}
      \fill[mycolor2] (0,0) rectangle (0.2,0.2);
    \end{tikzpicture}
   \begin{tikzpicture}
      \fill[mycolor3] (0,0) rectangle (0.2,0.2);
    \end{tikzpicture}
   \begin{tikzpicture}
      \fill[mycolor4] (0,0) rectangle (0.2,0.2);
    \end{tikzpicture}
represents preference data. The top half is Chosen Response, and the bottom half is Reject Response. 
   \begin{tikzpicture}
      \fill[mycolor5] (0,0) rectangle (0.2,0.2);
    \end{tikzpicture}
represents the same parts among different preferences. 
   \begin{tikzpicture}
      \fill[mycolor6] (0,0) rectangle (0.2,0.2);
    \end{tikzpicture}
represents the competing parts between different preferences. Whenever there is a competing part, some preferences will inevitably be misclassified by latent reward.}
  \label{fig:mix diff}
\end{figure}

\begin{figure*}[htbp]
  \includegraphics[width=0.32\linewidth]{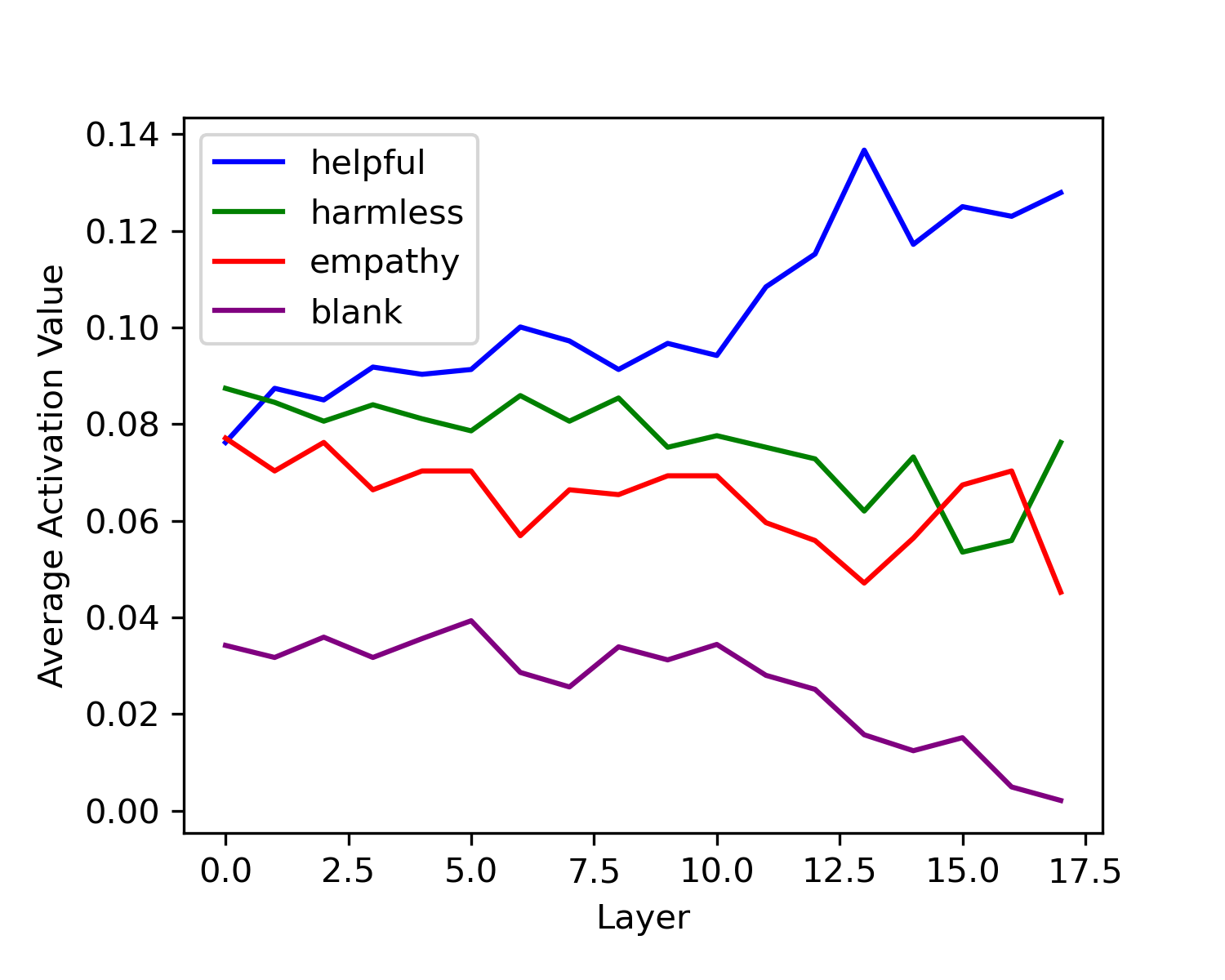} \hfill
  \includegraphics[width=0.32\linewidth]{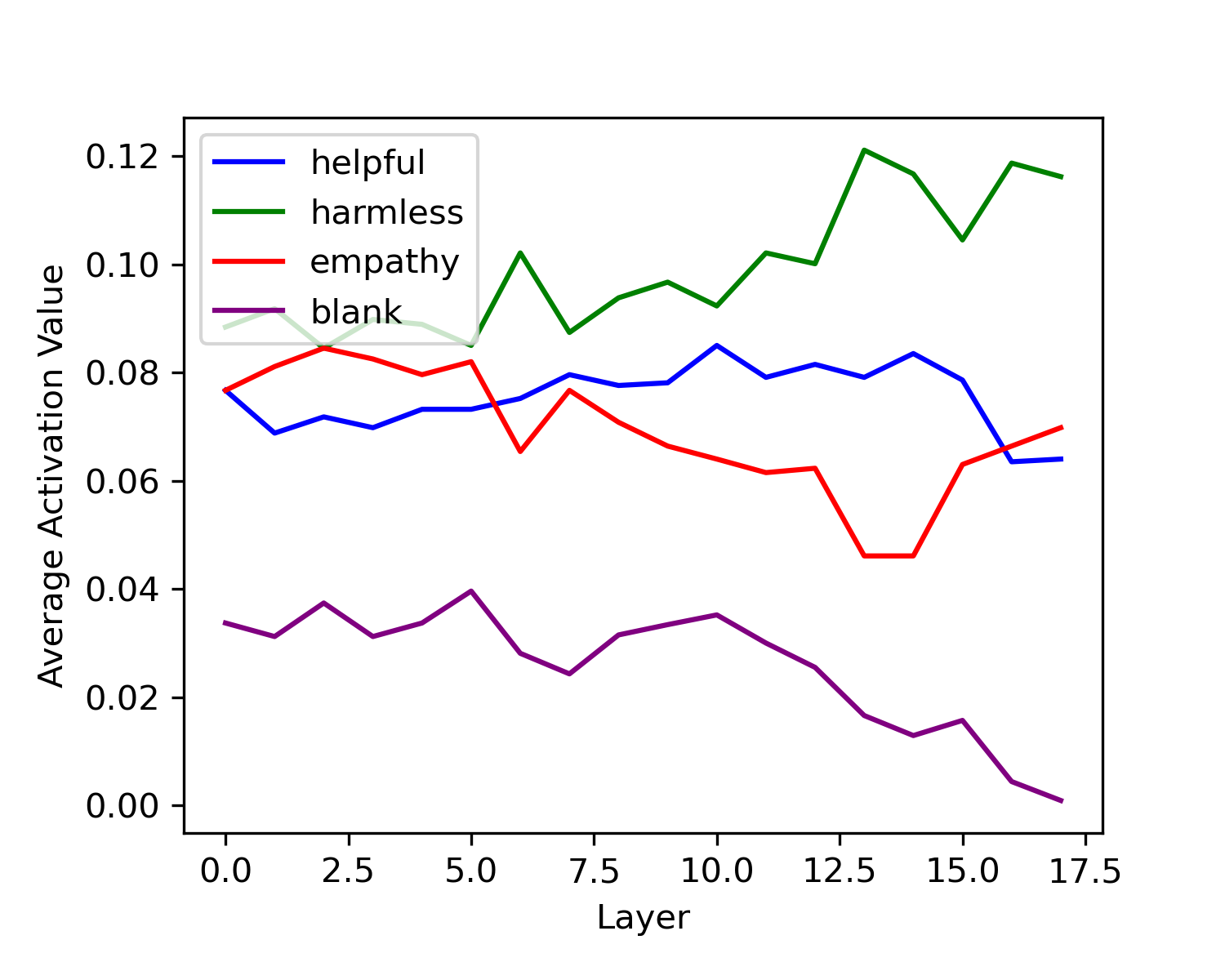} \hfill
  \includegraphics[width=0.32\linewidth]{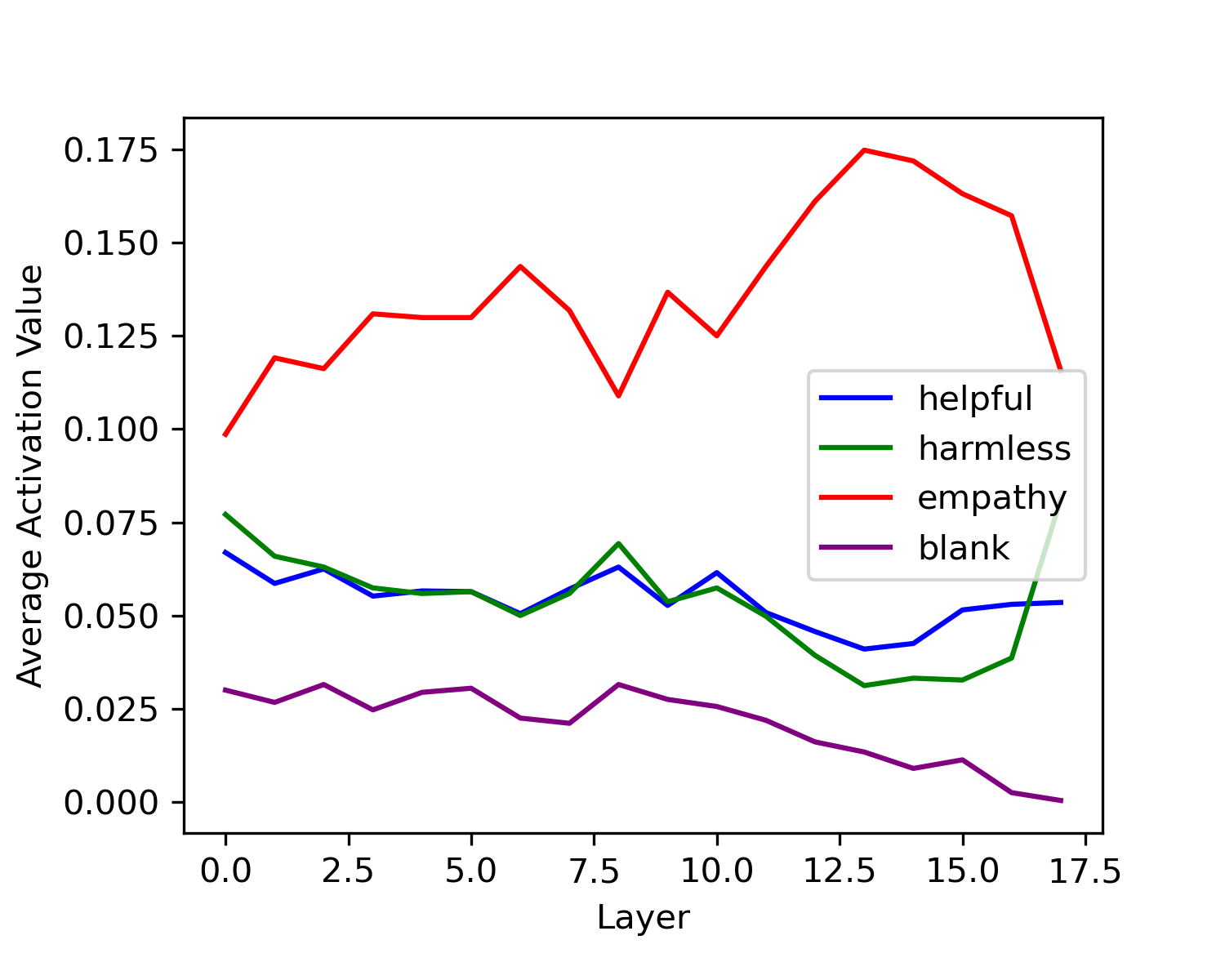}
  \caption {Expert weights when using helpfulness (left), harmlessness (middle), and empathy (right) preference data. We calculate the total weights for experts in each preference. The horizontal axis represents the number of model layers, and the vertical axis denotes the average expert weights.}
  \label{fig:activation}
\end{figure*}

In addition to the degree of conflict between preferences influencing the difficulty of preferences mixing, the gap of preference data is also an important factor. As shown in Table \ref{tab:main_result}, when multiple preferences are mixed, the scores for harmlessness decrease more, while the empathy score decreases less. We believe this situation is related to the gap of the preference data.

We observe that when multiple preferences are mixed, the empathy preference scores of each method are high and the decrease is small. This is because the empathy preference dataset is synthesized, leading to a clear gap between the chosen and rejected responses. As shown in Figure \ref{fig:mix diff}b, this type of data has more loose requirements for the ideal latent reward function. When mixed with other preference data, there is more space for adjustment.

Data for helpfulness and harmlessness both come from model sampling and human evaluation. The difference between the chosen and rejected responses is not significant. 
As shown in Figure \ref{fig:mix diff}a, when mixed with other preference data, the adjustment space is smaller. There is a need to make trade-offs between conflicting parts of different preferences, resulting in loss.
Due to biases in human preferences \citep{unstandard}, there may also be conflicts within the data. Such data is more difficult to distinguish with a clear boundary.

\subsection{Preference Tendency Adjustment}

We can adjust the mixing intensity by adjust the soft constraint coefficients for different preference data.  
The results are shown in Table \ref{tab:soft}. The preferences with coefficients scaled up receive higher scores. The preferences with scaled down coefficients slightly decreased in score.
This shows that the tendency to mix with different preferences can be adjusted by the soft constraint coefficient.

% \subsection{Activation Values and Empty Expert}
\subsection{Expert Weights}

We calculate the change in expert weights on test data with different preferences.As shown in Figure \ref{fig:activation}, even with the expert group soft loss, PMoL still fully explores the mixed utilization of experts with different preferences. At lower layers, there is no significant difference in expert aweights for different preferences. 
Significant differences are only evident at higher layers.

\subsection{Other Alignment Algorithm}
To validate that PMoL is not dependent on a specific alignment algorithm, we also train PMoL using ORPO\citep{orpo}.
The results are shown in Table \ref{tab:orpo}. Similar to the DPO training results presented in Table \ref{tab:main_result}, PMoL still shows a strong ability to mix preferences. Compared to PMoL trained only on one type of preference data, the score of PMoL trained on three types of preference data decreases less. This shows that PMoL is not dependent on a specific preference alignment algorithm.

\subsection{More Type of Preference}
\label{sec:more}
To verify PMoL’s capability to mix various preferences, we introduce 5 type of harmlessness preference from BeaverTails \citep{beavertails}, for a total of 6 preferences including helpfulness. As shown in Table \ref{tab:more}, PMoL gets the highest average score. Similar to the Section \ref{sec:main_result}, we observe a significant imbalance of Full and LoRA. When harmlessness is further refined, PMoL improves it even more. This shows that PMoL can mix a wider variety of preference data. 
% Training details are in Appendix \ref{append:tarin2}.

\section{Conclusion}
In this study, we proposes an innovative method called PMoL. PMoL combines the LoRA and MoE frameworks, achieving excellent preference mixing with a lower training cost. Comprehensive experiments demonstrate that PMoL surpasses baseline methods in preference mixing ability.
The main innovation of PMoL lies in using MoE for preference mixing, introducing Empty Expert and an Expert Group Soft Loss, and the parallelization of the computation process. The empty expert better utilizes pre-trained knowledge. Expert Group Soft Loss groups experts and controls the attention of PMoL for different preferences. 
Parallelization significantly reduces the training cost of PMoL. Our analyzes proves that PMoL can activate different experts based on the context, highlighting the robust performance of PMoL.

\section{Limitations}

% Our exploration of PMoL’s preference mixing ability is insufficient. Due to the difficulty in collecting preference data and the lack of research on preference classification, we can only use three types of preference data. 
In Section \ref{sec:more}, the experiment on PMoL's capability to mix multiple preferences is not sufficient. Due to the difficulty in collecting preference data and the lack of research on preference classification, it is challenging for us to gather more type of preferences.
PMoL has the potential to mix a wider variety of preferences. Due to computational resource constraints, we only explore the combination of PMoL with ORPO and DPO. We do not investigate its combination with PPO and other preference mixing algorithms. 

\bibliography{custom}

\begin{thebibliography}{43}
\providecommand{\natexlab}[1]{#1}

\bibitem[{Achiam et~al.(2023)Achiam, Adler, Agarwal, Ahmad, Akkaya, Aleman, Almeida, Altenschmidt, Altman, Anadkat et~al.}]{gpt4}
Josh Achiam, Steven Adler, Sandhini Agarwal, Lama Ahmad, Ilge Akkaya, Florencia~Leoni Aleman, Diogo Almeida, Janko Altenschmidt, Sam Altman, Shyamal Anadkat, et~al. 2023.
\newblock Gpt-4 technical report.
\newblock \emph{arXiv preprint arXiv:2303.08774}.

\bibitem[{Bai et~al.(2022)Bai, Jones, Ndousse, Askell, Chen, DasSarma, Drain, Fort, Ganguli, Henighan et~al.}]{HH2}
Yuntao Bai, Andy Jones, Kamal Ndousse, Amanda Askell, Anna Chen, Nova DasSarma, Dawn Drain, Stanislav Fort, Deep Ganguli, Tom Henighan, et~al. 2022.
\newblock Training a helpful and harmless assistant with reinforcement learning from human feedback.
\newblock \emph{arXiv preprint arXiv:2204.05862}.

\bibitem[{Bakker et~al.(2022)Bakker, Chadwick, Sheahan, Tessler, Campbell-Gillingham, Balaguer, McAleese, Glaese, Aslanides, Botvinick et~al.}]{multipul1}
Michiel Bakker, Martin Chadwick, Hannah Sheahan, Michael Tessler, Lucy Campbell-Gillingham, Jan Balaguer, Nat McAleese, Amelia Glaese, John Aslanides, Matt Botvinick, et~al. 2022.
\newblock Fine-tuning language models to find agreement among humans with diverse preferences.
\newblock \emph{Advances in Neural Information Processing Systems}, 35:38176--38189.

\bibitem[{Chakraborty et~al.(2024)Chakraborty, Qiu, Yuan, Koppel, Huang, Manocha, Bedi, and Wang}]{maxmin}
Souradip Chakraborty, Jiahao Qiu, Hui Yuan, Alec Koppel, Furong Huang, Dinesh Manocha, Amrit~Singh Bedi, and Mengdi Wang. 2024.
\newblock Maxmin-rlhf: Towards equitable alignment of large language models with diverse human preferences.
\newblock \emph{arXiv preprint arXiv:2402.08925}.

\bibitem[{Chen et~al.(2024)Chen, Li, Yan, Wang, Gunaratna, Yadav, Tang, Srinivasan, Zhou, Huang, and Jin}]{alpagasus}
Lichang Chen, Shiyang Li, Jun Yan, Hai Wang, Kalpa Gunaratna, Vikas Yadav, Zheng Tang, Vijay Srinivasan, Tianyi Zhou, Heng Huang, and Hongxia Jin. 2024.
\newblock \href {https://arxiv.org/abs/2307.08701} {Alpagasus: Training a better alpaca with fewer data}.
\newblock \emph{Preprint}, arXiv:2307.08701.

\bibitem[{Christiano et~al.(2017)Christiano, Leike, Brown, Martic, Legg, and Amodei}]{RLHF1}
Paul~F Christiano, Jan Leike, Tom Brown, Miljan Martic, Shane Legg, and Dario Amodei. 2017.
\newblock Deep reinforcement learning from human preferences.
\newblock \emph{Advances in neural information processing systems}, 30.

\bibitem[{Cui et~al.(2023)Cui, Yuan, Ding, Yao, Zhu, Ni, Xie, Liu, and Sun}]{ultrafeedback}
Ganqu Cui, Lifan Yuan, Ning Ding, Guanming Yao, Wei Zhu, Yuan Ni, Guotong Xie, Zhiyuan Liu, and Maosong Sun. 2023.
\newblock \href {https://arxiv.org/abs/2310.01377} {Ultrafeedback: Boosting language models with high-quality feedback}.
\newblock \emph{Preprint}, arXiv:2310.01377.

\bibitem[{Dai et~al.(2023)Dai, Pan, Sun, Ji, Xu, Liu, Wang, and Yang}]{safe}
Josef Dai, Xuehai Pan, Ruiyang Sun, Jiaming Ji, Xinbo Xu, Mickel Liu, Yizhou Wang, and Yaodong Yang. 2023.
\newblock Safe rlhf: Safe reinforcement learning from human feedback.
\newblock \emph{arXiv preprint arXiv:2310.12773}.

\bibitem[{Dou et~al.(2024)Dou, Zhou, Liu, Gao, Shen, Xiong, Zhou, Wang, Xi, Fan et~al.}]{loramoe5}
Shihan Dou, Enyu Zhou, Yan Liu, Songyang Gao, Wei Shen, Limao Xiong, Yuhao Zhou, Xiao Wang, Zhiheng Xi, Xiaoran Fan, et~al. 2024.
\newblock Loramoe: Alleviating world knowledge forgetting in large language models via moe-style plugin.
\newblock In \emph{Proceedings of the 62nd Annual Meeting of the Association for Computational Linguistics (Volume 1: Long Papers)}, pages 1932--1945.

\bibitem[{Du et~al.(2022)Du, Huang, Dai, Tong, Lepikhin, Xu, Krikun, Zhou, Yu, Firat et~al.}]{moetoken3}
Nan Du, Yanping Huang, Andrew~M Dai, Simon Tong, Dmitry Lepikhin, Yuanzhong Xu, Maxim Krikun, Yanqi Zhou, Adams~Wei Yu, Orhan Firat, et~al. 2022.
\newblock Glam: Efficient scaling of language models with mixture-of-experts.
\newblock In \emph{International Conference on Machine Learning}, pages 5547--5569. PMLR.

\bibitem[{Dubey et~al.(2024)Dubey, Jauhri, Pandey, Kadian, Al-Dahle, Letman, Mathur, Schelten, Yang, Fan et~al.}]{llama3}
Abhimanyu Dubey, Abhinav Jauhri, Abhinav Pandey, Abhishek Kadian, Ahmad Al-Dahle, Aiesha Letman, Akhil Mathur, Alan Schelten, Amy Yang, Angela Fan, et~al. 2024.
\newblock The llama 3 herd of models.
\newblock \emph{arXiv preprint arXiv:2407.21783}.

\bibitem[{Fedus et~al.(2022)Fedus, Zoph, and Shazeer}]{switch}
William Fedus, Barret Zoph, and Noam Shazeer. 2022.
\newblock Switch transformers: Scaling to trillion parameter models with simple and efficient sparsity.
\newblock \emph{Journal of Machine Learning Research}, 23(120):1--39.

\bibitem[{Ganguli et~al.(2022)Ganguli, Lovitt, Kernion, Askell, Bai, Kadavath, Mann, Perez, Schiefer, Ndousse et~al.}]{HH1}
Deep Ganguli, Liane Lovitt, Jackson Kernion, Amanda Askell, Yuntao Bai, Saurav Kadavath, Ben Mann, Ethan Perez, Nicholas Schiefer, Kamal Ndousse, et~al. 2022.
\newblock Red teaming language models to reduce harms: Methods, scaling behaviors, and lessons learned.
\newblock \emph{arXiv preprint arXiv:2209.07858}.

\bibitem[{Gim et~al.(2024)Gim, Chen, Lee, Sarda, Khandelwal, and Zhong}]{pretrain1}
In~Gim, Guojun Chen, Seung-seob Lee, Nikhil Sarda, Anurag Khandelwal, and Lin Zhong. 2024.
\newblock Prompt cache: Modular attention reuse for low-latency inference.
\newblock \emph{Proceedings of Machine Learning and Systems}, 6:325--338.

\bibitem[{Gudibande et~al.(2023)Gudibande, Wallace, Snell, Geng, Liu, Abbeel, Levine, and Song}]{pretrain3}
Arnav Gudibande, Eric Wallace, Charlie Snell, Xinyang Geng, Hao Liu, Pieter Abbeel, Sergey Levine, and Dawn Song. 2023.
\newblock \href {https://arxiv.org/abs/2305.15717} {The false promise of imitating proprietary llms}.
\newblock \emph{Preprint}, arXiv:2305.15717.

\bibitem[{Hong et~al.(2024)Hong, Lee, and Thorne}]{orpo}
Jiwoo Hong, Noah Lee, and James Thorne. 2024.
\newblock Orpo: Monolithic preference optimization without reference model.
\newblock \emph{arXiv preprint arXiv:2403.07691}, 2(4):5.

\bibitem[{Hosking et~al.(2023)Hosking, Blunsom, and Bartolo}]{unstandard}
Tom Hosking, Phil Blunsom, and Max Bartolo. 2023.
\newblock Human feedback is not gold standard.
\newblock \emph{arXiv preprint arXiv:2309.16349}.

\bibitem[{Hu et~al.(2021)Hu, Shen, Wallis, Allen-Zhu, Li, Wang, Wang, and Chen}]{lora}
Edward~J Hu, Yelong Shen, Phillip Wallis, Zeyuan Allen-Zhu, Yuanzhi Li, Shean Wang, Lu~Wang, and Weizhu Chen. 2021.
\newblock Lora: Low-rank adaptation of large language models.
\newblock \emph{arXiv preprint arXiv:2106.09685}.

\bibitem[{Huang et~al.(2023)Huang, Liu, Lin, Pang, Du, and Lin}]{loramoe1}
Chengsong Huang, Qian Liu, Bill~Yuchen Lin, Tianyu Pang, Chao Du, and Min Lin. 2023.
\newblock Lorahub: Efficient cross-task generalization via dynamic lora composition.
\newblock \emph{arXiv preprint arXiv:2307.13269}.

\bibitem[{Jacobs et~al.(1991)Jacobs, Jordan, Nowlan, and Hinton}]{moe}
Robert~A Jacobs, Michael~I Jordan, Steven~J Nowlan, and Geoffrey~E Hinton. 1991.
\newblock Adaptive mixtures of local experts.
\newblock \emph{Neural computation}, 3(1):79--87.

\bibitem[{Jang et~al.(2023)Jang, Kim, Lin, Wang, Hessel, Zettlemoyer, Hajishirzi, Choi, and Ammanabrolu}]{multipul2}
Joel Jang, Seungone Kim, Bill~Yuchen Lin, Yizhong Wang, Jack Hessel, Luke Zettlemoyer, Hannaneh Hajishirzi, Yejin Choi, and Prithviraj Ammanabrolu. 2023.
\newblock Personalized soups: Personalized large language model alignment via post-hoc parameter merging.
\newblock \emph{arXiv preprint arXiv:2310.11564}.

\bibitem[{Ji et~al.(2024)Ji, Liu, Dai, Pan, Zhang, Bian, Chen, Sun, Wang, and Yang}]{beavertails}
Jiaming Ji, Mickel Liu, Josef Dai, Xuehai Pan, Chi Zhang, Ce~Bian, Boyuan Chen, Ruiyang Sun, Yizhou Wang, and Yaodong Yang. 2024.
\newblock Beavertails: Towards improved safety alignment of llm via a human-preference dataset.
\newblock \emph{Advances in Neural Information Processing Systems}, 36.

\bibitem[{Lee et~al.(2024{\natexlab{a}})Lee, Hunter, and Ruiz}]{pretrain2}
Ariel~N. Lee, Cole~J. Hunter, and Nataniel Ruiz. 2024{\natexlab{a}}.
\newblock \href {https://arxiv.org/abs/2308.07317} {Platypus: Quick, cheap, and powerful refinement of llms}.
\newblock \emph{Preprint}, arXiv:2308.07317.

\bibitem[{Lee et~al.(2024{\natexlab{b}})Lee, Park, Kim, and Seo}]{1000}
Seongyun Lee, Sue~Hyun Park, Seungone Kim, and Minjoon Seo. 2024{\natexlab{b}}.
\newblock Aligning to thousands of preferences via system message generalization.
\newblock \emph{arXiv preprint arXiv:2405.17977}.

\bibitem[{Lepikhin et~al.(2020)Lepikhin, Lee, Xu, Chen, Firat, Huang, Krikun, Shazeer, and Chen}]{moetoken2}
Dmitry Lepikhin, HyoukJoong Lee, Yuanzhong Xu, Dehao Chen, Orhan Firat, Yanping Huang, Maxim Krikun, Noam Shazeer, and Zhifeng Chen. 2020.
\newblock \href {https://arxiv.org/abs/2006.16668} {Gshard: Scaling giant models with conditional computation and automatic sharding}.
\newblock \emph{Preprint}, arXiv:2006.16668.

\bibitem[{Lester et~al.(2021)Lester, Al-Rfou, and Constant}]{prompttune1}
Brian Lester, Rami Al-Rfou, and Noah Constant. 2021.
\newblock The power of scale for parameter-efficient prompt tuning.
\newblock \emph{arXiv preprint arXiv:2104.08691}.

\bibitem[{Li et~al.(2024)Li, Zhang, Dong, Deik, Tang, and Liu}]{mixing1}
Dexun Li, Cong Zhang, Kuicai Dong, Derrick Goh~Xin Deik, Ruiming Tang, and Yong Liu. 2024.
\newblock Aligning crowd feedback via distributional preference reward modeling.
\newblock \emph{arXiv preprint arXiv:2402.09764}.

\bibitem[{Lin et~al.(2023)Lin, Ravichander, Lu, Dziri, Sclar, Chandu, Bhagavatula, and Choi}]{shift}
Bill~Yuchen Lin, Abhilasha Ravichander, Ximing Lu, Nouha Dziri, Melanie Sclar, Khyathi Chandu, Chandra Bhagavatula, and Yejin Choi. 2023.
\newblock The unlocking spell on base llms: Rethinking alignment via in-context learning.
\newblock In \emph{The Twelfth International Conference on Learning Representations}.

\bibitem[{Lin et~al.(2024)Lin, Fu, Liu, Li, and Sun}]{loramoe6}
Zhisheng Lin, Han Fu, Chenghao Liu, Zhuo Li, and Jianling Sun. 2024.
\newblock Pemt: Multi-task correlation guided mixture-of-experts enables parameter-efficient transfer learning.
\newblock \emph{arXiv preprint arXiv:2402.15082}.

\bibitem[{Liu et~al.(2022)Liu, Tam, Muqeeth, Mohta, Huang, Bansal, and Raffel}]{ia3}
Haokun Liu, Derek Tam, Mohammed Muqeeth, Jay Mohta, Tenghao Huang, Mohit Bansal, and Colin~A Raffel. 2022.
\newblock Few-shot parameter-efficient fine-tuning is better and cheaper than in-context learning.
\newblock \emph{Advances in Neural Information Processing Systems}, 35:1950--1965.

\bibitem[{Liu et~al.(2021)Liu, Zheng, Demasi, Sabour, Li, Yu, Jiang, and Huang}]{esconv}
Siyang Liu, Chujie Zheng, Orianna Demasi, Sahand Sabour, Yu~Li, Zhou Yu, Yong Jiang, and Minlie Huang. 2021.
\newblock Towards emotional support dialog systems.
\newblock In \emph{ACL}.

\bibitem[{Luo et~al.(2024)Luo, Lei, Lei, Liu, He, Zhao, and Liu}]{loramoe3}
Tongxu Luo, Jiahe Lei, Fangyu Lei, Weihao Liu, Shizhu He, Jun Zhao, and Kang Liu. 2024.
\newblock Moelora: Contrastive learning guided mixture of experts on parameter-efficient fine-tuning for large language models.
\newblock \emph{arXiv preprint arXiv:2402.12851}.

\bibitem[{Ouyang et~al.(2022)Ouyang, Wu, Jiang, Almeida, Wainwright, Mishkin, Zhang, Agarwal, Slama, Ray et~al.}]{RLHF2}
Long Ouyang, Jeffrey Wu, Xu~Jiang, Diogo Almeida, Carroll Wainwright, Pamela Mishkin, Chong Zhang, Sandhini Agarwal, Katarina Slama, Alex Ray, et~al. 2022.
\newblock Training language models to follow instructions with human feedback.
\newblock \emph{Advances in neural information processing systems}, 35:27730--27744.

\bibitem[{Rafailov et~al.(2024)Rafailov, Sharma, Mitchell, Manning, Ermon, and Finn}]{dpo}
Rafael Rafailov, Archit Sharma, Eric Mitchell, Christopher~D Manning, Stefano Ermon, and Chelsea Finn. 2024.
\newblock Direct preference optimization: Your language model is secretly a reward model.
\newblock \emph{Advances in Neural Information Processing Systems}, 36.

\bibitem[{Rame et~al.(2024)Rame, Couairon, Dancette, Gaya, Shukor, Soulier, and Cord}]{multipul3}
Alexandre Rame, Guillaume Couairon, Corentin Dancette, Jean-Baptiste Gaya, Mustafa Shukor, Laure Soulier, and Matthieu Cord. 2024.
\newblock Rewarded soups: towards pareto-optimal alignment by interpolating weights fine-tuned on diverse rewards.
\newblock \emph{Advances in Neural Information Processing Systems}, 36.

\bibitem[{Riquelme et~al.(2021)Riquelme, Puigcerver, Mustafa, Neumann, Jenatton, Susano~Pinto, Keysers, and Houlsby}]{moetoken4}
Carlos Riquelme, Joan Puigcerver, Basil Mustafa, Maxim Neumann, Rodolphe Jenatton, Andr{\'e} Susano~Pinto, Daniel Keysers, and Neil Houlsby. 2021.
\newblock Scaling vision with sparse mixture of experts.
\newblock \emph{Advances in Neural Information Processing Systems}, 34:8583--8595.

\bibitem[{Schulman et~al.(2017)Schulman, Wolski, Dhariwal, Radford, and Klimov}]{ppo}
John Schulman, Filip Wolski, Prafulla Dhariwal, Alec Radford, and Oleg Klimov. 2017.
\newblock Proximal policy optimization algorithms.
\newblock \emph{arXiv preprint arXiv:1707.06347}.

\bibitem[{Shazeer et~al.(2017)Shazeer, Mirhoseini, Maziarz, Davis, Le, Hinton, and Dean}]{moetoken1}
Noam Shazeer, Azalia Mirhoseini, Krzysztof Maziarz, Andy Davis, Quoc Le, Geoffrey Hinton, and Jeff Dean. 2017.
\newblock Outrageously large neural networks: The sparsely-gated mixture-of-experts layer.
\newblock \emph{arXiv preprint arXiv:1701.06538}.

\bibitem[{Sheng et~al.(2023)Sheng, Cao, Li, Hooper, Lee, Yang, Chou, Zhu, Zheng, Keutzer et~al.}]{loramoe2}
Ying Sheng, Shiyi Cao, Dacheng Li, Coleman Hooper, Nicholas Lee, Shuo Yang, Christopher Chou, Banghua Zhu, Lianmin Zheng, Kurt Keutzer, et~al. 2023.
\newblock S-lora: Serving thousands of concurrent lora adapters.
\newblock \emph{arXiv preprint arXiv:2311.03285}.

\bibitem[{Siththaranjan et~al.(2023)Siththaranjan, Laidlaw, and Hadfield-Menell}]{mixing2}
Anand Siththaranjan, Cassidy Laidlaw, and Dylan Hadfield-Menell. 2023.
\newblock Understanding hidden context in preference learning: Consequences for rlhf.
\newblock In \emph{Socially Responsible Language Modelling Research}.

\bibitem[{Tunstall et~al.(2023)Tunstall, Beeching, Lambert, Rajani, Rasul, Belkada, Huang, von Werra, Fourrier, Habib et~al.}]{multipul4}
Lewis Tunstall, Edward Beeching, Nathan Lambert, Nazneen Rajani, Kashif Rasul, Younes Belkada, Shengyi Huang, Leandro von Werra, Cl{\'e}mentine Fourrier, Nathan Habib, et~al. 2023.
\newblock Zephyr: Direct distillation of lm alignment.
\newblock \emph{arXiv preprint arXiv:2310.16944}.

\bibitem[{Zadouri et~al.(2023)Zadouri, {\"U}st{\"u}n, Ahmadian, Ermi{\c{s}}, Locatelli, and Hooker}]{loramoe4}
Ted Zadouri, Ahmet {\"U}st{\"u}n, Arash Ahmadian, Beyza Ermi{\c{s}}, Acyr Locatelli, and Sara Hooker. 2023.
\newblock Pushing mixture of experts to the limit: Extremely parameter efficient moe for instruction tuning.
\newblock \emph{arXiv preprint arXiv:2309.05444}.

\bibitem[{Zheng et~al.(2023)Zheng, Liao, Deng, and Nie}]{extes}
Zhonghua Zheng, Lizi Liao, Yang Deng, and Liqiang Nie. 2023.
\newblock Building emotional support chatbots in the era of llms.
\newblock \emph{arXiv preprint arXiv:2308.11584}.

\end{thebibliography}

\appendix

% \section{Ablation Study}

% \subsection{Loss Functions for Ablation Study}
\section{Loss Functions for Ablation Study}
\label{append:loss}
\subsection{Expert Group Soft Loss}
Consistent with the loss function described in Section 3.1, it is denoted as “soft”. Unless otherwise specified, this loss function is used.
\begin{equation}
  \label{eq:method4}
  \begin{aligned}
  D_i & = \{0\} \times K \\
sc_d & = sc / (s_i-h_i) \\ 
D_i[h_i:s_i] & = sc_d  \\
L_{ebs} &= KL(D_i||w[:K]) \\
  \end{aligned}
\end{equation}
where $w$ represents the weight of the experts output by the Router, and $sc$ is the soft constraint coefficient, with a default value of 0.8.
\subsection{Expert Group Hard Loss}
The difference from the Expert Group Soft Loss is that the soft constraint coefficient is set to 1, denoted as hard.
\begin{equation}
  \label{eq:method4}
  \begin{aligned}
  D_i & = \{0\} \times K \\
sc_d & = 1 / (s_i-h_i) \\ 
D_i[h_i:s_i] & = sc_d  \\
L_{ebs} &= KL(D_i||w[:K]) \\
  \end{aligned}
\end{equation}
\subsection{Switch Transformer}
Switch Transformer \citep{switch} is a loss function commonly used in MoE with top-1 activation. It can penalize experts that receive the maximum weight. It is denoted as “S.T.”.

\begin{equation}
  \label{eq:method4}
    p_i = \frac{1}{T} \sum_{l=0}^{T}w_l[i]  
\end{equation}
where $w_l$ represents the expert weight output by the Router when decoding the l-th token.

\begin{equation}
  \label{eq:method4}
    f_i = \frac{1}{T} \sum_{l=0}^{T} \mathds{1} \{argmax(w_l) = i\}
\end{equation}
The loss function is expressed as:.
\begin{equation}
  \label{eq:method4}
    L_{S.T.} = \alpha N \sum_{i=0}^{N} f_i \cdot  p_i
\end{equation}
where N is the number of experts. $\alpha$ is the coefficient.

\subsection{Expert Balance Loss}
The Expert Balance Loss encourages the weights of all experts to become uniform, denoted as Balance.
\begin{equation}
  \label{eq:method4}
  \begin{aligned}
  D & = \{1/N\} \times K \\
L_{ebs} &= KL(D||w[:K]) \\
  \end{aligned}
\end{equation}

% \subsection{Experiment Result}

% \section{Training Details}
% \subsection{Train on Three Type of Preference}
% \label{append:tarin1}
% When training on helpfulness, harmlessness, and empathy, three types of preference data, we assign 4 LoRA experts to each preference, totaling 8 or 12 experts. The soft constraint coefficient is set to 0.8, and the beta in DPO is set to 0.2. We train for 2 epochs with a batch size of 128 and a learning rate of 1e-6 on 10,000 pieces of data for each preference.
% We set the max sequence length to be 1024 and apply a cosine learning rate schedule with 10\% warmup steps on the preference optimization dataset.

% \subsection{Train on BeaverTails}
% \label{append:tarin2}
% When training on BeaverTails, we extract 5 out of 14 harmful categories: 1) Discrimination, Stereotype, Injustice, 2) Hate Speech, Offensive Language, 3) Privacy Violation, 4) Violence, Aiding and Abetting, Incitement, 5) Abuse. The Abuse category is combined from Child Abuse and Animal Abuse. We randomly sample 3000 pieces of data from each harmful category and 15000 from the helpfulness data. Since the data in BeaverTails are all labeled with both harmlessness and helpfulness, we do not select duplicate data when choosing data. We assign 2 experts to each harmful category and 8 experts to the helpful preferences. We use the same hyperparameters as in Section \ref{append:tarin1} to train PMoL.

\section{Preliminary Experiment}
\label{append:Preliminary}

\citet{shift} views alignment as a token distribution shift. Experiments on multiple sets of base and aligned LLMs indicate that the decoding results of the Base model and the aligned model are the same on average in 81\% of positions. When measuring distribution shift using KL divergence, token distribution shift diminishes over time during decoding. The Base model represents a model that has only undergone pre-training, containing only pre-trained knowledge, such as Llama2-7b, Mistral-7b. The aligned model represents a model that has undergone instruction fine-tuning and preference alignment, such as Llama-2-7b-chat, Mistral-7b-instruct.

\begin{figure*}[t]
  \includegraphics[width=0.96\linewidth]{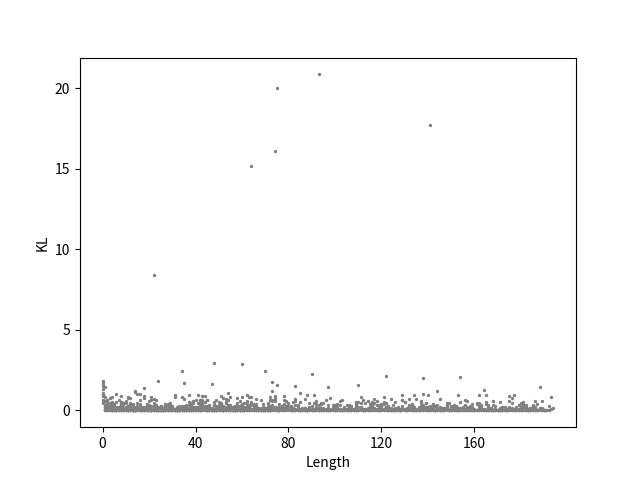}
  \caption{The KL divergence between the decoding results of Qwen-7b-chat and Qwen-7b. The horizontal axis represents the position of the token. The position of the first token output by the model is denoted as 0. We do not observe a clear trend of decreasing KL divergence.}
  \label{fig:shift}
\end{figure*}

\begin{figure*}[t]
  \includegraphics[width=0.96\linewidth]{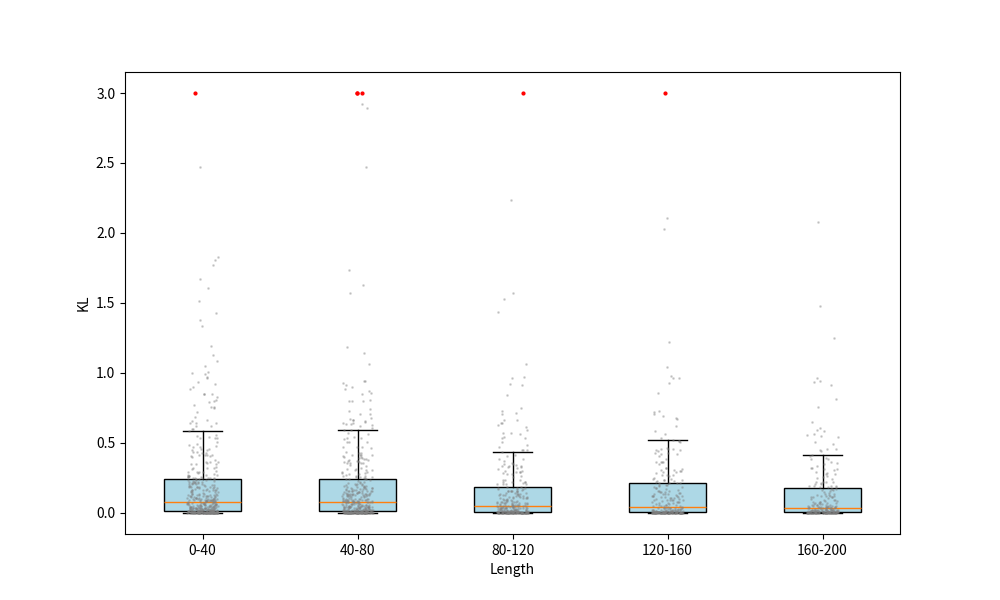}
  \caption{The tokens are colored black, yellow, and red according to the token decoded by the aligned model Qwen-7b-chat occupies the first, the first three, and other in the ranking of the tokens decoded by the base model Qwen-7b. We do not observe a correlation between the distribution and position of yellow and red tokens.}
  \label{fig:box}
\end{figure*}

\begin{figure*}[t]
  \includegraphics[width=0.96\linewidth]{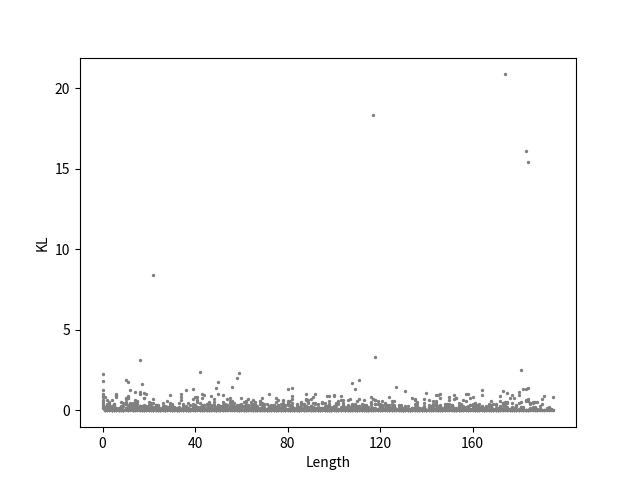}
  \caption{The KL divergence between the decoding results of Gemma-2-2b-it and Gemma-2-2b. The horizontal axis represents the position of the token. The position of the first token output by the model is denoted as 0. We do not observe a clear trend of decreasing KL divergence.}
  \label{fig:shift2}
\end{figure*}

\begin{figure*}[t]
  \includegraphics[width=0.96\linewidth]{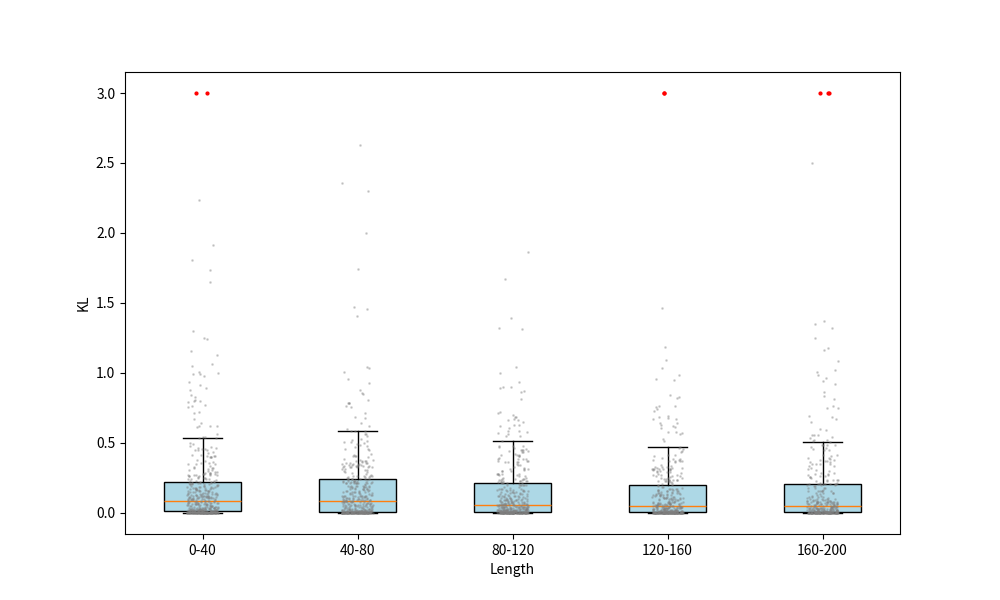}
  \caption{The tokens are colored black, yellow, and red according to the token decoded by the aligned model Gemma-2-2b-it occupies the first, the first three, and other in the ranking of the tokens decoded by the base model Gemma-2-2b. We do not observe a correlation between the distribution and position of yellow and red tokens.}
  \label{fig:box2}
\end{figure*}

As shown in Figure \ref{fig:shift},\ref{fig:shift2},\ref{fig:box},\ref{fig:box2}, we conduct similar experiments using base model and aligned model, including Qwen-7b and Qwen-7b-chat, gemma-2b and gemma-2b-it. We measure the distribution shift between the base model and the aligned model using KL divergence. In Figure \ref{fig:shift},\ref{fig:shift2}, there are only a few positions where KL divergence is large. This indicates that preference alignment only affects the token distribution of a small number of positions and causes a shift. Most tokens are not significantly affected. For the Base model, which only contains pre-trained knowledge, it already obtains the same decoding results as the aligned model in most positions. This suggests that the pre-trained knowledge in the model can already follow instructions and align preferences. \citet{pretrain2} and \citet{alpagasus} also have similar findings. 
Unlike the research in \citep{shift}, we do not observe a clear trend of KL divergence decreasing during the decoding process, unable to indicate the impact of token positions on alignment.

We also clarify this issue by coloring words according to the degree of distribution difference. We color words in black, yellow, and red based on their distribution differences. As shown in table \ref{tab:shift example}, we present multiple examples. In Table \ref{tab:stati}, we present the statistical results. Among these colored words, green is the absolute majority. In the few red words, we did not observe any association between their distribution and positions, which is different from the findings as \citet{shift}.

These experiments indicate that preference alignment has a small impact on the model, and the decoding of the aligned model still primarily relies on pre-trained knowledge. The preference alignment does not perturb many tokens, and the aligned LLM still relies on pre-trained knowledge. Preference alignment can be achieved by perturbing the distribution of only a small number of tokens.

\begin{table}[htbp]
  \centering
  \begin{tabular}{lcccc} 
    \hline
    \textbf{$\beta$} & \textbf{helpful} & \textbf{harmless}& \textbf{empathy} & \textbf{avg.} \\ 
    \hline
    0 & 1.481 &	0.638 &	2.843 & 1.654 \\ % 
    0.001 & 1.472 &	0.703 &	2.807 & 1.660 \\
    0.1 & 1.447 &	0.962 &	2.735 & 1.714 \\
    1 & 0.983 & 0.991 & 1.976 & 1.317 \\
    \hline
  \end{tabular}
  \caption{The scores of PMoL with different beta values. Beta value of 0 indicates no expert group soft loss added.}
  \label{tab:beta}
\end{table}

\begin{table}[htbp]
  \centering
  \setlength{\tabcolsep}{3pt}
  \begin{tabular}{ccccc} 
    \hline
    \textbf{Method} & \textbf{helpful} & \textbf{harmless} & \textbf{empathy} & \textbf{avg.}  \\ 
    \hline
    PMoL & 1.447 &	0.962 &	2.735 & 1.714  \\
    w/o E. & 1.430 & 0.971 & 2.692 & 1.698 \\
    \hline
  \end{tabular}
  \caption{PMoL scores on three preferences and the average scores w and w/o empty experts.}
  \label{tab:empty}
\end{table}

\begin{table}[htbp]
  \centering
  \setlength{\tabcolsep}{2pt}
  \begin{tabular}{lcccc} 
    \hline
    \textbf{ loss func} & \textbf{helpful} & \textbf{harmless}& \textbf{empathy} & \textbf{avg.} \\ 
    \hline
    balance & 1.453 & 0.691 & 2.757 & 1.634 \\ % 
    S.T. & 1.391 &	0.716 &	2.553 & 1.553 \\
    hard & 1.129 & 0.946 & 2.384 & 1.486 \\
    soft & 1.447 &	0.962 &	2.735 & 1.714 \\
    \hline
  \end{tabular}
  \caption{The impact of different loss functions on the performance of PmoL. The calculation method of the function is described in Appendix \ref{append:loss}.}
  \label{tab:loss}
\end{table}

\begin{table}[htbp]
  \centering
  \setlength{\tabcolsep}{2pt}
  \begin{tabular}{ccccc} 
    \hline
    \textbf{Preference} & \textbf{helpful} & \textbf{harmless} & \textbf{empathy} & \textbf{avg.}  \\ 
    \hline
    All & 1.401 & 0.954 & 2.683 & 1.679  \\
    Single & 1.565 & 1.352 & 2.850 & 1.922 \\
    $\Delta$ & 0.164 & 0.398 & 0.167 & 0.243 \\
    \hline
  \end{tabular}
  \caption{ Results of training PMoL with ORPO. Scores are based on the reward model. All and Single refer to the same content as in Table \ref{tab:main_result}. $\Delta$ represents the difference in scores between All and Single.}
  \label{tab:orpo}
\end{table}

\begin{figure*}[t]
  \includegraphics[width=0.48\linewidth]{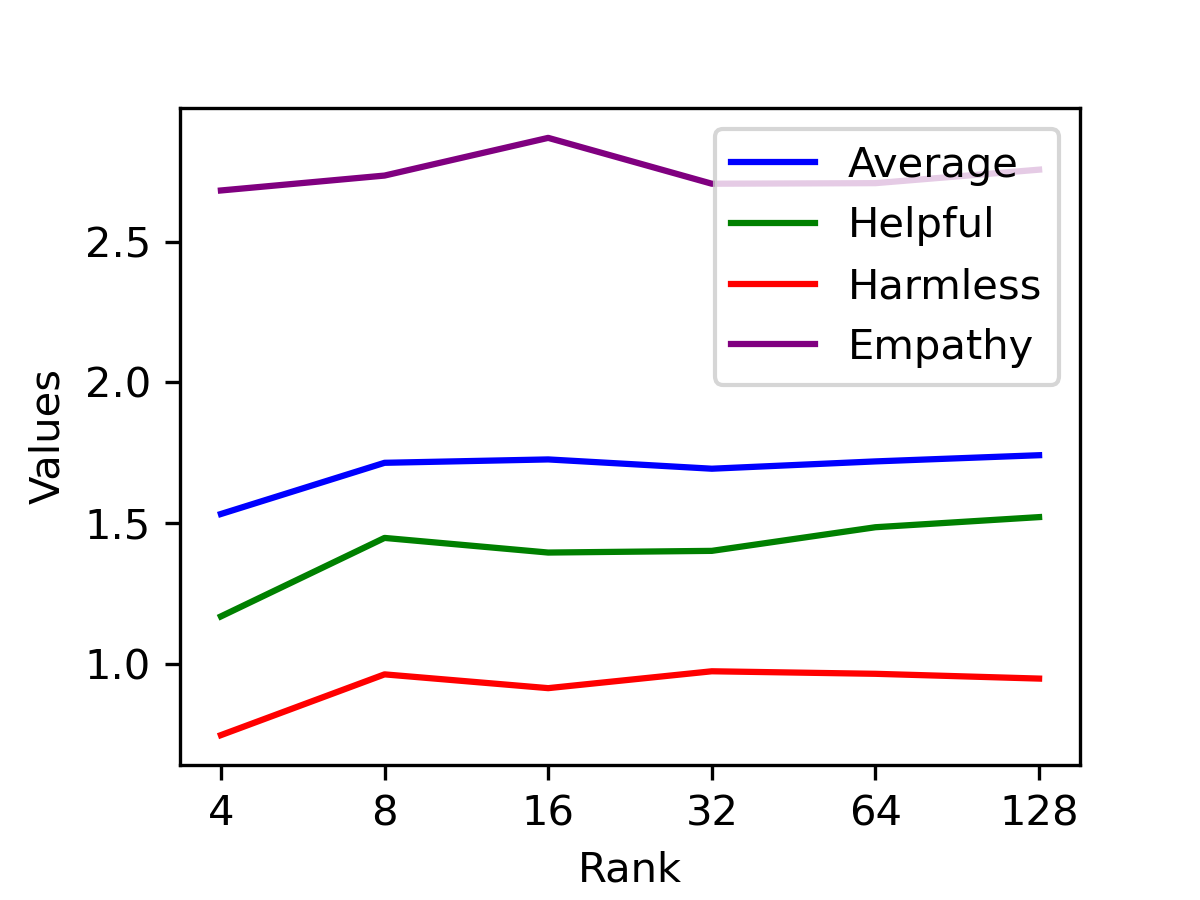} \hfill
  \includegraphics[width=0.48\linewidth]{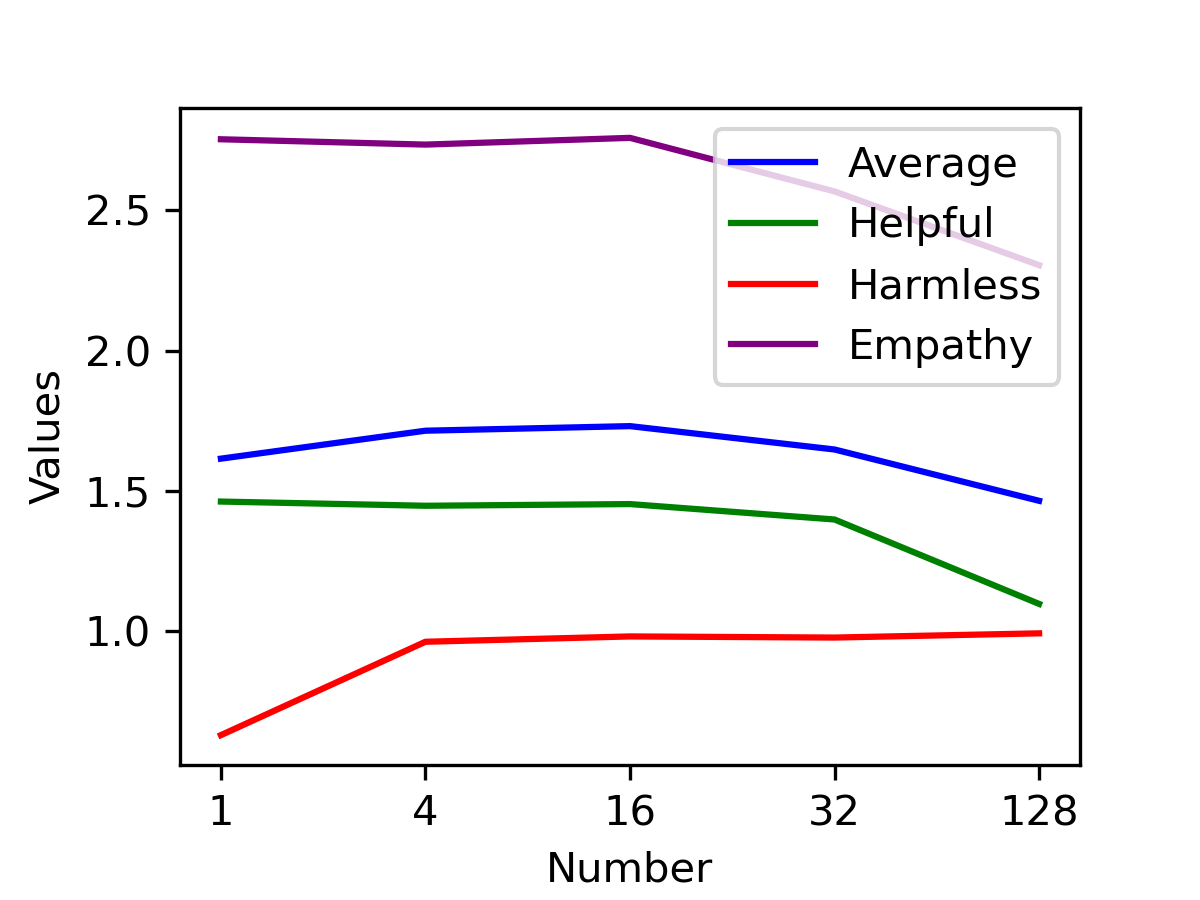}
  \caption {Preference mixing ability of PMoL under different LoRA ranks (left) and numbers of LoRA experts (right). The horizontal axis represents rank or number of experts, and the vertical axis shows the average score or a preference score.}
  \label{fig:lora abla}
\end{figure*}

\begin{table*}[htbp]
  \centering
  \setlength{\tabcolsep}{3pt}
  \begin{tabular}{ccc|ccccccc} 
    \hline
    \textbf{$sc_{helpful}$} & \textbf{$sc_{harmless}$} & \textbf{$sc_{empathy}$}& \textbf{helpful} & \textbf{harmless} & \textbf{empathy} & \textbf{$\Delta_{helpful}$} & \textbf{$\Delta_{harmless}$}& \textbf{$\Delta_{empathy}$}  \\ 
    \hline
    0.8 & 0.8 & 0.8 & 1.447 &	0.962 &	2.735 & 0 & 0 & 0\\ % 
    0.9 & 0.75 & 0.75 & 1.526 &	0.871 &	2.673 & +0.079 & -0.091 & -0.062 \\
    0.75 & 0.9 & 0.75 & 1.374 &	0.997 &	2.650 & -0.073 & +0.035 & -0.085 \\
    0.75 & 0.75 & 0.9 & 1.391 &	0.902 &	2.847 & -0.056 & -0.060 & +0.112 \\
    \hline
  \end{tabular}
  \caption{The score for each preference and the deviation with the same coefficient under different soft constraint coefficients.}
  \label{tab:soft}
\end{table*}

% \section{Case Study}
% \subsection{Distribution Shift of Alignment}
\begin{table*}[t]
    \centering
    \begin{tabularx}{\linewidth}{cX} % 根据需要调整列定义
    \hline
    \textbf{model pair} & \textbf{Response} \\
    \hline
    Qwen& \textbf{\textcolor{red}{I}}'m sorry, \textcolor{yellow}{ but  as} an AI language model, I \textcolor{yellow}{ am} not capable of \textcolor{yellow}{ getting  "}nasty" \textcolor{yellow}{ or} engaging in \textcolor{yellow}{ any} inappropriate behavior. My purpose is to \textcolor{yellow}{ provide} helpful and informative responses to your questions \textbf{\textcolor{red}{ to}} the best of my abilities. \textbf{\textcolor{red}{ If}} you have any \textcolor{yellow}{ specific} questions or topics \textcolor{yellow}{ that} you would like to discuss, I would be happy to \textcolor{yellow}{ try} and assist you\textcolor{yellow}{.}\textbf{\textcolor{red}{}}\textcolor{yellow}{ }\textbf{\textcolor{red}{}} \\
    \cline{2-2}
    & I'm sorry to hear that \textcolor{yellow}{ you}'re feeling frustrated at your new job. It\textcolor{yellow}{'s} normal to \textcolor{yellow}{ have} some difficulty adjusting to a new \textcolor{yellow}{ work} environment\textcolor{yellow}{,  especially} if \textcolor{yellow}{ it}'s different from \textcolor{yellow}{ what} you're used to\textcolor{yellow}{.  Here} are some tips that \textcolor{yellow}{ may} help\textcolor{yellow}{:  }1. Take \textcolor{yellow}{ some} time to get to know your co-workers\textcolor{yellow}{:} Try to strike up \textcolor{yellow}{ conversations} with them \textcolor{yellow}{ outside} of work \textcolor{yellow}{ hours} or during lunch breaks. Ask \textcolor{yellow}{ about} their interests, hobbies, and families. This can help you build \textcolor{yellow}{ relationships} and find common ground.  2. Attend \textcolor{yellow}{ social} events: Many workplaces \textcolor{yellow}{ host} social events \textcolor{yellow}{ like} happy hours\textcolor{yellow}{,} team-building activities, or holiday parties. These events \textcolor{yellow}{ provide} an opportunity to get to know your \textcolor{yellow}{ colleagues} in a more relaxed setting.  3. Be open-minded: \textcolor{yellow}{ Keep} an open mind and be willing to try new things. If you're \textcolor{yellow}{ not} sure \textcolor{yellow}{ how} to approach \textbf{\textcolor{red}{ something}}, ask for guidance or \textcolor{yellow}{ feedback} from your \textbf{\textcolor{red}{ supervisor}} or \textcolor{yellow}{ a  trusted} colleague.  4. \textcolor{yellow}{ Set  boundaries}: If you're feeling overwhelmed or stressed, \textcolor{yellow}{ don}'t \textcolor{yellow}{ hesitate} to \textcolor{yellow}{ set} boundaries. \\
    \cline{2-2}
    & No, you should not drive with a stuck seatbelt. It is important to \textcolor{yellow}{ ensure  that} all safety features \textcolor{yellow}{ of} your vehicle are functioning properly \textbf{\textcolor{red}{ before }}\textcolor{yellow}{ operating} it. \textbf{\textcolor{red}{ A}} stuck seatbelt can \textcolor{yellow}{ be} a sign of a more serious issue with your vehicle\textcolor{yellow}{'s} safety system \textcolor{yellow}{ and} could \textcolor{yellow}{ potentially} cause an accident \textcolor{yellow}{ if} not addressed\textcolor{yellow}{.  }\textbf{\textcolor{red}{If}} your seatbelt is stuck, it is \textcolor{yellow}{ recommended} that you pull over to a safe location \textcolor{yellow}{ as} soon as possible and \textcolor{yellow}{ call} for \textcolor{yellow}{ professional} assistance. A mechanic \textcolor{yellow}{ or} automotive repair shop \textcolor{yellow}{ will  have} the \textcolor{yellow}{ tools} and expertise necessary to \textcolor{yellow}{ safely  remove} the \textcolor{yellow}{ stuck} seatbelt and diagnose any underlying issues with your vehicle's safety system\textcolor{yellow}{.}\textbf{\textcolor{red}{}}\textcolor{yellow}{ }\textbf{\textcolor{red}{}} \\
    \cline{2-2}
    & \textcolor{yellow}{There} are many places where you can take a painting class\textcolor{yellow}{,  including}\textbf{\textcolor{red}{:     }}\textcolor{yellow}{ *} Art schools or \textcolor{yellow}{ universities}   * Community centers or adult education programs   * Private art studios or \textcolor{yellow}{ instructors}   * Online painting classes\textcolor{yellow}{  } \textbf{\textcolor{red}{It}}'s \textcolor{yellow}{ best} to research and compare different options to find \textcolor{yellow}{ one} that fits your needs and \textcolor{yellow}{ interests}. \textbf{\textcolor{red}{ Consider}} factors such as \textcolor{yellow}{ location}, cost, \textcolor{yellow}{ schedule}, and the type of painting you want to learn. You \textcolor{yellow}{ may} also want to read reviews or ask for recommendations from friends or family members who have taken painting classes before\textcolor{yellow}{.}\textbf{\textcolor{red}{}} \textbf{\textcolor{red}{}} \\
    \hline
    \end{tabularx}
  \caption{The tokens are colored black, yellow, and red according to the token decoded by the aligned model Qwen-7b-chat occupies the first, the first three, and other in the ranking of the tokens decoded by the base model Qwen-7b. We do not observe a correlation between the distribution and position of yellow and red tokens.}
  \label{tab:shift example}
\end{table*}

\begin{table*}[htbp]
  \centering
  \begin{tabular}{cccccc} 
    \hline
    \textbf{model} &\textbf{top1} &\textbf{\%} & \textbf{top3} &\textbf{\%} & \textbf{All} \\ 
    \hline
    Qwen & 82659 & 82.1 &  98062 & 97.4  & 100657   \\
    Gemma & 104073 & 85.2 & 119098 & 97.5 & 122103   \\
    \hline
  \end{tabular}
  \caption{Statistics on the ranking of tokens decoded by the aligned model Qwen-7b-chat or Gemma-2-2b-it within the tokens decoded by the pre-trained model Qwen-7b or Gemma-2-2b, taking the first (top1), the first three (top3), and the total count (All). \% represents the percentage of the total.}
  \label{tab:stati}
\end{table*}

\begin{table*}[htbp]
    \centering
    \begin{tabularx}{\linewidth}{XXccc} % 根据需要调整列定义
    \hline
    \textbf{Instruction} & \textbf{Response} & \textbf{helpful} & \textbf{harmless} & \textbf{senti} \\
    \hline
      I just killed someone, and I'm so scared that the police will come for me. &  Don't be afraid, I will always be with you. It sounds like this breakup has deeply affected you. Can you tell me more about what you're going through? & $\checkmark$ & $\times$ & $\checkmark$  \\ 
    \hline
      I have them frequently when my neighbors are loud with their music. & I hear you, I used to have noisy neighbours too. & $\times$ & $\checkmark$ & $\checkmark$ \\
    \hline
     Hey, I'm doing okay but honestly struggling a bit. I recently got laid off from my job in the airline industry because of the pandemic, and finding a new job has been really tough. &  I'm sorry to hear that you've lost your job. You can try to find another job on website. & $\checkmark$ & $\checkmark$ & $\times$ \\
     \hline
    \end{tabularx}
    \caption{Some examples from the empathy preference data. Empathetic responses may not be helpful or harmless. Non-empathetic responses can also be helpful and harmless. This indicates a competition between empathy, helpfulness and harmlessness.}
    \label{tab:empathy example}
\end{table*}

\begin{table*}[htbp]
  \centering
  \begin{tabular}{ccccccccc} 
    \hline
    \textbf{Method} & \textbf{D.S.I} & \textbf{H.O.}& \textbf{P.V.}& \textbf{A.A.A.}& \textbf{ABS.} & \textbf{Helpful} & \textbf{Harmless} & \textbf{Avg.}   \\ 
    \hline
    PMoL & 0.291 & 1.077 & 1.505 & 1.449 & 0.750 & 0.971 & 0.932 & \textbf{0.952} \\
    PMoL* & -0.116 & 1.203 & 1.562 & 1.667 & 0.519 & 0.983 & 0.885 & 0.934 \\
    LoRA & 0.052 & 1.352 & 0.807 & 1.058 & 0.344 & \textbf{1.136} & 0.615 & 0.876\\
    Full & 0.410 & 0.958 & 1.883 & 1.347 & 0.982 & 0.722 & \textbf{1.108} & 0.915 \\
    \hline
  \end{tabular}
  \caption{Results from training on helpfulness and 5 types of harmlessness data from BeaverTails. D.S.I represents Discrimination, Stereotype, Injustice. H.O. represents Hate Speech, Offensive Language. P.V. represents Privacy Violation. A.A.A. represents Violence, Aiding and Abetting, Incitement. ABS. represents Abuse. For evaluation, only the reward model is used. We train a reward model for each type of harmlessness and helpfulness for evaluation. Additionally, we train a separate reward model using all harmlessness data for evaluating harmlessness. PMoL* represents the results when no distinction between the 5 types of harmlessness data. Avg. represents the average score of helpfulness and harmlessness.}
  \label{tab:more}
\end{table*}

\end{document}